\documentclass[journal]{IEEEtran}
\newcommand{\sysName}{SketchDesc}
\newcommand{\pixelLevel}{pixel-level }
\usepackage{diagbox} %
\usepackage{multirow}
\newcommand{\tabincell}[2]{\begin{tabular}{@{}#1@{}}#2\end{tabular}}
\usepackage{amsmath}
\usepackage{amsfonts}
\usepackage{bm}

\ifCLASSOPTIONcompsoc
  \usepackage[nocompress]{cite}
\else
  \usepackage{cite}
\fi

\ifCLASSINFOpdf

  \usepackage[pdftex]{graphicx}
  \graphicspath{{../pdf/}{../jpeg/}}
  \DeclareGraphicsExtensions{.pdf,.jpeg,.png}
\else
  \usepackage[dvips]{graphicx}
  \graphicspath{{../eps/}}
  \DeclareGraphicsExtensions{.eps}
\fi

\begin{document}
\title{SketchDesc: Learning Local Sketch Descriptors for Multi-view Correspondence}

\author{Deng~Yu,
        Lei~Li,
        Youyi~Zheng,
        Manfred~Lau,
        Yi-Zhe~Song,~\IEEEmembership{Senior Member,~IEEE},
        Chiew-Lan~Tai,
        Hongbo~Fu\textsuperscript{$\dagger$}
\thanks{This work was partially supported by grants from 
the Research Grants Council of the Hong Kong Special Administrative Region, China (Project No. CityU 11206319, CityU 11212119, CityU 11237116, HKUST16210718), and the City University of Hong Kong (Project No. 7005176 (SCM)). Youyi Zheng was partially supported by the China Young 1000 Talent Program and the Fundamental Research Funds for the Central Universities. 
}
\thanks{D. Yu, M. Lau, and H. Fu are with the School of Creative Media, City University of Hong Kong. H. Fu is the corresponding author of this paper. 
E-mail: \{deng.yu@my., manfred.lau@, hongbofu@\}cityu.edu.hk}
\thanks{ L. Li and C.-L. Tai are with the Department of Computer Science and
Engineering, Hong Kong University of Science and Technology. 
E-mail: \{llibb, taicl\}@cse.ust.hk}
\thanks{ Y. Zheng is with the State Key Lab of CAD\&CG, Zhejiang University. 
E-mail: youyizheng@zju.edu.cn}
\thanks{ Y.-Z. Song is with SketchX, CVSSP, University of Surrey. 
E-mail: y.song@surrey.ac.uk}
\thanks{Copyright © 20xx IEEE. Personal use of this material is permitted. However, permission to use this material for any other purposes must be obtained from the IEEE by sending an email to pubs-permissions@ieee.org.}
}
\maketitle


\begin{abstract}
In this paper, we study the problem of multi-view sketch correspondence, where we take as input multiple freehand sketches with different views of the same object and predict as output the
semantic correspondence among the sketches.
This problem is challenging since the visual features of corresponding points at different views can be very different. To this end,
we take a deep learning approach and learn a novel local sketch descriptor
from data. We contribute a training dataset by generating
the \pixelLevel correspondence for the multi-view line drawings 
synthesized from 3D shapes. 
To handle the sparsity and ambiguity of sketches,
we design a novel 
multi-branch neural network 
that integrates a patch-based representation and a multi-scale strategy to learn the \pixelLevel correspondence among multi-view sketches. 
We demonstrate the effectiveness of our proposed approach with extensive experiments on hand-drawn sketches
and  multi-view line drawings rendered from multiple 3D shape datasets.
\end{abstract}

\begin{IEEEkeywords}
  Multi-view sketches,
  correspondence learning, multi-scale,  
  patch-based descriptor.
\end{IEEEkeywords}
\IEEEpeerreviewmaketitle


\section{Introduction}
\begin{figure}[t]
    \begin{center}
    \includegraphics[width=1\linewidth]{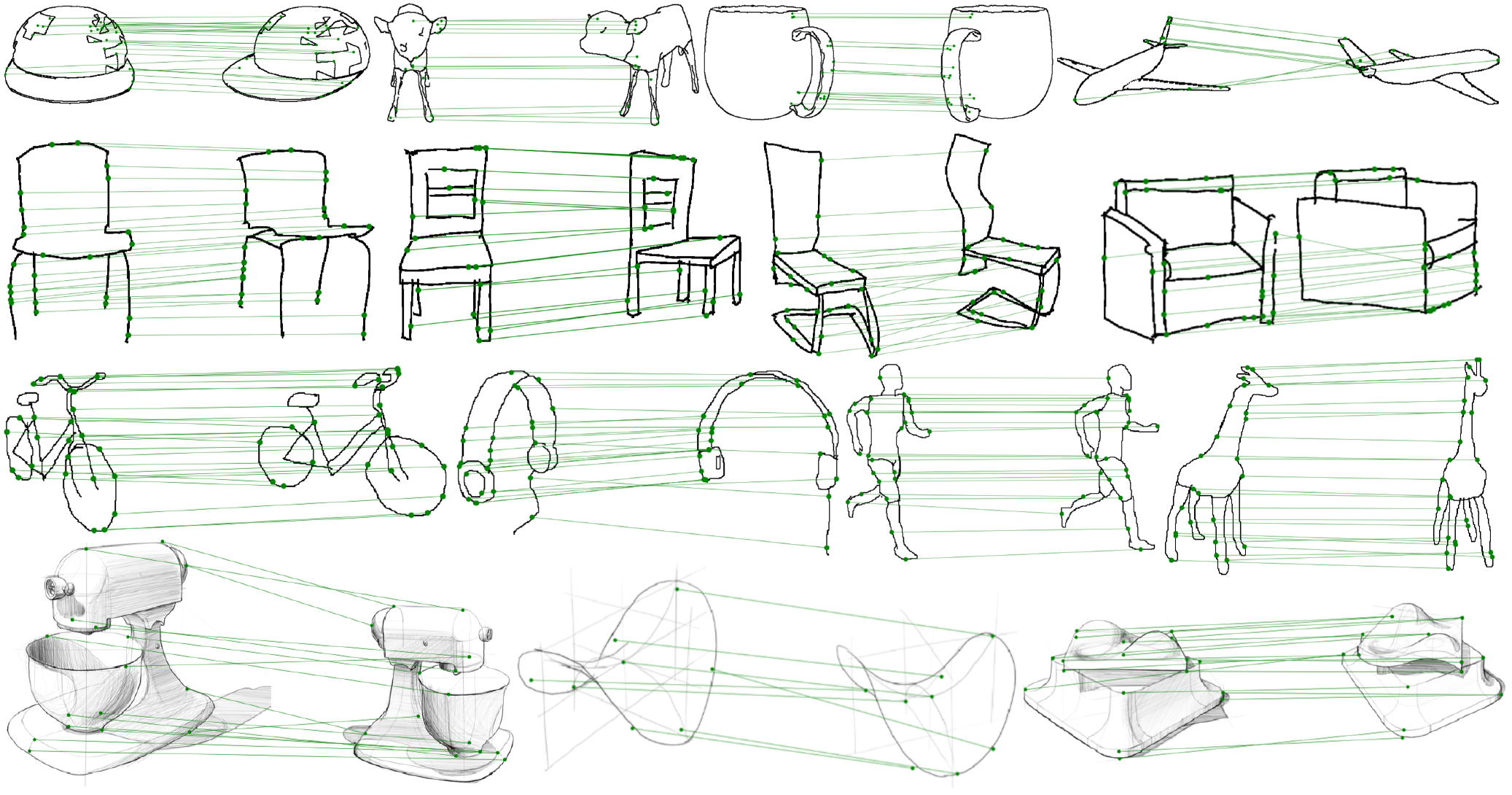}
    \end{center}
       \caption{
       Multi-view sketch correspondence results by \sysName~on line drawings synthesized from 3D shapes (top) and freehand sketches (middle and bottom). 
       }
    \label{fig:Figure matchingReults}
 \end{figure}

Sketching as a universal form of communication provides arguably the most natural and direct way for humans to render and interpret the visual world. 
While it is not challenging for human viewers to interpret missing 3D information from single-view sketches, multi-view inputs are often needed for computer algorithms to recover the underlying 3D geometry due to the inherent ambiguity in single-view sketches. A key problem for interpreting multi-view sketches of the same object is to establish semantic correspondence among them. This problem has been mainly studied for 3D geometry reconstruction from careful engineering drawings in orthographic views \cite{governi20133d}.
The problem of establishing semantic correspondence from rough sketches in arbitrary views (e.g., those in Figure \ref{fig:Figure matchingReults} ({middle and bottom}))
is challenging due to the abstraction and distortions in both shape and view, and is largely unexplored. Addressing this problem can benefit various applications, for example to design an interactive interface for users with little training in drawing to create 3D shapes using multi-view sketches.

In this work, we take the first step to learn to establish the semantic correspondence between freehand sketches
depicting the same object from different views. 
This demands a proper shape descriptor.
However, traditional descriptors like Shape Context \cite{belongie2001shape} and recent learning-based patch descriptors \cite{tian2017l2,mishchuk2017working,tian2019sosnet} are often designed to be invariant to 2D transformations (e.g. rotation, translation, and limited distortion) and cannot handle large view changes (e.g. with view disparity greater than 30 degrees).
Such descriptors, especially used for the applications of stereo matching \cite{hirschmuller2008evaluation,tombari2008classification,chen2015deep,luo2016efficient,yang2019hierarchical} and image-based 3D modeling \cite{tung2009complete,liu2010ray},
heavily exploit the features containing textures and shadings of images for inferring similarities among different points or image patches, and thus are not directly applicable to our problem. This is because sketches only contain binary lines and points, exhibiting inherent sparsity and ambiguity.

We observe that human viewers can easily identify corresponding points from sketches with very different views. This is largely because human viewers have knowledge of sketched objects in different views. Our idea is thus to adapt deep neural networks previously used for learning patch-based descriptors to learn descriptors for corresponding patches from multi-view sketches. 

Training deep neural networks require a 
large-scale dataset
of sketch images with ground truth semantic correspondence. Unfortunately, such training datasets are not available. 
On the other hand, manually collecting multi-view hand-drawn sketches and labeling the ground-truth correspondence would be a demanding task.
Following previous deep learning solutions for 3D interpretation of sketches
~\cite{nishida2016interactive,delanoy20183d,li2018robust}, we synthesize the multi-view line drawings from 3D shapes (e.g. from ShapeNet \cite{yi2016scalable}) by using non-photorealistic rendering. 
Given the synthesized dataset of multi-view line drawings as sketches, we project the 3D vertices of 3D models to multi-view sketches to get ground-truth correspondences 
(Figure \ref{fig:Figure dataPineline}). 
This data generation pipeline is to emphasize the correspondence from 3D shapes and force deep networks to learn the valuable 3D correspondences among 2D multi-view sketches.

We formulate the correspondence learning problem into a metric function learning procedure and build upon the latest techniques for metric learning and descriptor learning \cite{tian2017l2}. 
To find an effective feature descriptor for multi-view sketches, we combine a
patch-based representation and a multi-scale strategy (Figure \ref{fig:Figure TripLets}) 
to address the abstraction problem of sketches (akin to Sketch-A-Net \cite{yu2017sketch}).  

We further design a multi-branch network (Figure \ref{fig:Figure Network}) with shared-weights to process the patches at different scales.
The patch-based representation helps the network specify the local features of a point on a sketch image by embedding the information of its neighboring pixels. Our multi-scale strategy feeds the network with the local and global perspectives to learn the distinctive information
at different scales.

The multi-scale patch representation allows the use of ground-truth correspondences that are away from sketch lines as the additional training data. This not only improves the correspondence accuracy but also enables correspondences inside the regions of sketched objects (e.g., the cup in Figure \ref{fig:Figure matchingReults}), potentially benefiting applications like sketch-based 3D shape synthesis.
We evaluate our method by performing multi-view sketch correspondence 
and 
pixel-wise retrieval tasks on a large-scale dataset of synthesized multi-view sketches based on three shape repositories: ShapeNet, Princeton Segmentation Benchmark (PSB) \cite{chen2009benchmark} and Structure Recovery \cite{shen2012structure} to show the effectiveness of our proposed framework.
We also test our trained network on freehand sketches (Figure~\ref{fig:Figure matchingReults}) drawn by volunteers and collected from the OpenSketch dataset \cite{gryaditskaya2019opensketch}, which shows its robustness against shape and view distortions. 

Our contributions are summarized as follows: 
\begin{itemize}
    \item To the best of our knowledge, we are among the first to study the problem of multi-view sketch correspondence.
    \item We introduce a multi-branch neural network
    to learn a patch-based descriptor that can be used to measure semantic similarity between patches from multi-view sketches. We demonstrate the improved accuracy for sketch correspondence achieved by using our descriptor compared to other existing descriptors.
    \item We collect a large-scale dataset consisting of 
    6,852 multi-view sketches with ground-truth correspondences of 587 shapes spanning 18 object categories, and will release the dataset to the research community.
\end{itemize}
\section{Related Work}
\label{section:Related Work}
We review the literature that is closely related to our work, namely, the approaches for correspondence establishment in images, and the approaches for sketch analysis using deep learning technologies.

\subsection{Correspondence Establishment
for Images}

\textit{Image-based Modeling and Stereo Matching.} 
Image-based modeling often takes as input multiple images of an object \cite{liu2010ray,agarwal2011building} or scene \cite{tung2009complete,snavely2006photo} from different views, and aims to reconstruct the underlying 3D geometry.
A typical approach to this problem is to first detect a sparse set of key points,
then adopt a feature descriptor (e.g., SIFT \cite{lowe2004distinctive}) to describe the patches centered at the key points, and finally conduct feature matching to build the correspondence among multi-view images.
Stereo matching \cite{hirschmuller2008evaluation,tombari2008classification} takes two images from different but often close viewpoints and aims to establish dense pixel-level correspondence across images.
Our problem is different from these tasks in the following ways. First, the view disparity in our input sketches is often much larger. Second, unlike natural photos, which have rich textures, our sketches have more limited information due to their line-based representation.

\textit{Local Image Descriptors.}
Local image descriptors are typically derived from image patches centered at points of interest and designed to be invariant to certain factors, such as rotation, scale, or intensity, for robustness.
Existing local image descriptors can be broadly categorized as hand-crafted descriptors and learning-based descriptors.
A full review is beyond the scope of this work.
We refer the interested readers to~\cite{Csurka:2018:FHDLIF} for an insightful survey.

Classical descriptors include, to name a few, SIFT \cite{lowe2004distinctive}, SURF \cite{bay2006surf}, Shape Context \cite{belongie2001shape}, and HOG \cite{dalal2005histograms}.
The conventional local descriptors are mostly built upon low-level image properties and constructed using hand-crafted rules.
Recently, learning-based local descriptors produced by deep convolutional neural networks (CNNs) \cite{tian2017l2,mishchuk2017working,tian2019sosnet} have shown their superior performance over the hand-crafted descriptors, owing importantly to the availability of large-scale image correspondence datasets~\cite{WB07:BrownDataset,Balntas_2017_CVPR} obtained from 3D reconstructions.
To learn robust 2D local descriptors, extensive research has been dedicated to the development of CNN designs~\cite{Han:2015:MatchNet,Zagoruyko:2015:LCIP,tian2017l2,Zhang_2017_ICCV,Wei_2018_CVPR}, loss functions~\cite{G_2016_CVPR,Song_2016_CVPR,Balntas:2016:LLFD,mishchuk2017working,Keller_2018_CVPR,mishkin2018repeatability} and training strategies~\cite{Simo:2015:DLDCFPD,choy_nips16,luo2018geodesc}.
The above methods, however, are not specially designed for learning multi-view sketch correspondence. The work of GeoDesec \cite{luo2018geodesc} shares the closest spirit to ours and employs geometry constraints from 3D reconstruction by Structure-from-Motion (SfM) to refine the training data. 
However, SfM heavily depends on the textures and shadings in the image domain and is not suitable for sketches.

\begin{figure}[t]
   \begin{center}
   \includegraphics[width=1\linewidth]{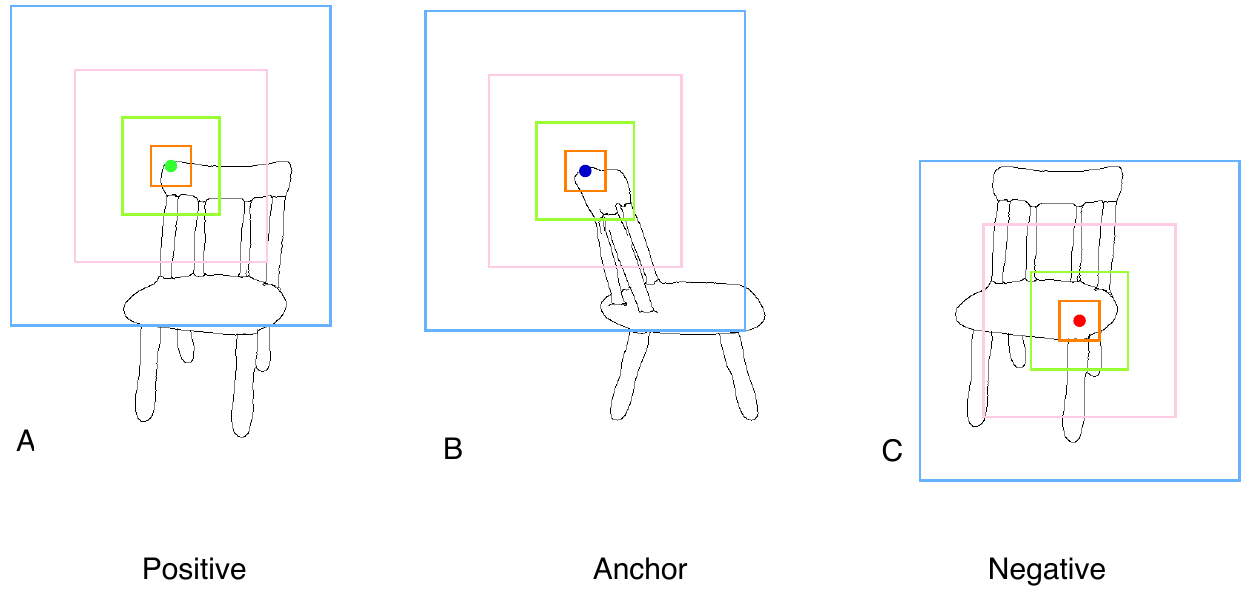}
   \end{center}
      \caption{Illustration of our multi-scale patch-based representation.  
   Given a 480$\times$480 sketch input and a pixel of interest,
   we use multi-scale (32$\times$32, 64$\times$64, 128$\times$128, and  256$\times$256) patches to capture the neighborhood of the pixel.
	The positive, anchor and negative patches are formed as a training triplet.} 
   \label{fig:Figure TripLets}
   \end{figure}

\textit{Multi-scale Strategy for Descriptors.}
Shilane and Funkhouser \cite{shilane2007distinctive} computed a 128-D descriptor at four scales 
in spherical regions to describe distinctive regions on 3D shape surfaces. 
Recently, Huang et al. \cite{huang2018learning} proposed to learn 3D point descriptors from multi-view projections with progressively zoomed out viewpoints.
Inspired by these methods, we design a multi-scale strategy to gather local and global context to locate corresponding points in sketches across views (Figure \ref{fig:Figure TripLets}). 
Different from \cite{huang2018learning}, which focuses on rendered patches of 3D shapes,
our work considers patches of sparse line drawings with limited textures as input.
To reduce the network size and computation,
we adopt a smaller input scale of $32\times32$~\cite{tian2017l2} rather than $224\times224$ in \cite{huang2018learning}. 
In addition, Huang et al. \cite{huang2018learning} used three viewpoints for 3D point descriptors, while our work extracts descriptors of points in sketches drawn under a specific viewpoint for correspondence establishment across a larger range of views.

\begin{figure}
   \begin{center}
   \includegraphics[width=0.6\linewidth]{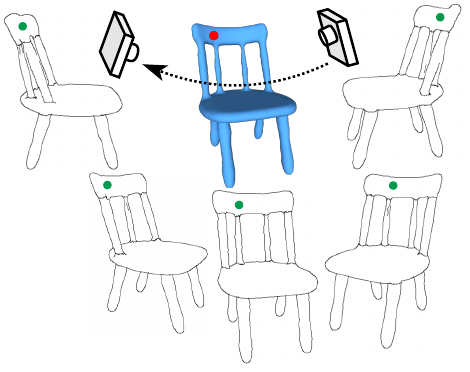}
   \end{center}
      \caption{
      Illustration of synthesized multi-view sketches (of size 480 $\times$ 480), with corresponding pixels (in green) projected from the same vertex (in red) on a 3D shape.
      }
   \label{fig:Figure dataPineline}
\end{figure}

\subsection{Deep Learning in Sketch Analysis}
With the recent advances in deep learning techniques, a variety of deep learning based methods have been proposed for sketch analysis tasks such as sketch synthesis \cite{ha2017neural},  {face sketch-photo synthesis \cite{wang2014comprehensive,zhu2019deep,zhu2019face}}, sketch recognition \cite{yu2017sketch}, sketch segmentation \cite{li2018sketch}, and sketch retrieval \cite{sangkloy2016sketchy,xu2018sketchmate}. Among these works, \cite{zhu2019deep} and \cite{zhu2019face} are the most relevant to our work, and they aim 
to model the correspondence between face photos and face sketches. However, different from 
our task to exploit the \pixelLevel correspondence between sketches, they explore the image-level mapping between face photos and sketches. In general, none of these existing solutions can be directly applied to our task.
To make deep learning possible in sketch analysis,
there exists multiple large-scale datasets of sketches including the TU-Berlin \cite{eitz2012humans}, QuickDraw \cite{ha2017neural}, and Sketchy \cite{sangkloy2016sketchy} datasets. However, they do not contain multi-view sketches and thus cannot be used to train our network.
A sketch is generally represented as either a rasterized binary-pixel image \cite{sangkloy2016sketchy} or vector sequences \cite{ha2017neural,li2018sketch} or both \cite{xu2018sketchmate}.
Since it is difficult to render line drawings from 3D shapes as a sequence of well-defined strokes, we use rasterized binary-pixel images to represent our training data and input sketches.

\textit{Multi-view Sketch Analysis.}
Multi-view sketches are often used in sketch-based 3D modeling \cite{rivers20103d,nishida2016interactive,lun20173d,delanoy20183d,li2018robust}. 
Early sketch-based modeling methods (e.g., \cite{rivers20103d}) require precise engineering drawings as input 
or are limited by demanding mental efforts, requiring users to first decompose a desired 3D shape into parts and then constructing each part through careful engineering drawings \cite{governi20133d}. 
To alleviate this issue, several recent methods \cite{nishida2016interactive,lun20173d,delanoy20183d,li2018robust} leverage  learning-based frameworks (e.g., GAN \cite{Goodfellow2014Generative}) 
to obtain the priors from training data and then infer 3D shapes from novel input multi-view sketches (usually
in orthographic views, namely, the front, back, and side views). 
However, these methods often process individual multi-view sketches in separate branches and do not explicitly consider the semantic correspondence between input sketches.  
Recently, a richly-annotated dataset of product design sketches, named OpenSketch, was presented by 
\cite{gryaditskaya2019opensketch},
which contains around 400 sketches of 12 man-made objects. 
However, the limited data in OpenSketch is insufficient for training deep neural networks. 
In our work, we synthesized 6,852 sketches for 18 shape categories to learn local sketch descriptors.

SketchZooms by Navarro et al. \cite{navarro2019sketchzooms} is a concurrent work and studies the sketch correspondence problem with a similar deep learning based solution. Our work is different from SketchZooms as follows.
To generate training data for cross-object correspondence, the used 3D shapes need to be semantically registered together in advance in SketchZooms,
while our work is more fine-grained as it explores training data generation from individual 3D shapes for cross-view correspondence.
SketchZooms follows \cite{huang2018learning,su2015multi} and adapts AlexNet~\cite{AlexNet:NIPS2012_4824} (40M parameters) with the final layer replaced with a view pooling layer.
In contrast,  our designed framework has a smaller size (1.4M parameters), and its high performance 
shown in our experiments paves a way for our method to be more easily integrated into mobile or touch devices.

\section{Methodology}
\label{section:Methodology}

\begin{figure*}[t]
\begin{center}
\includegraphics[width=1\linewidth]{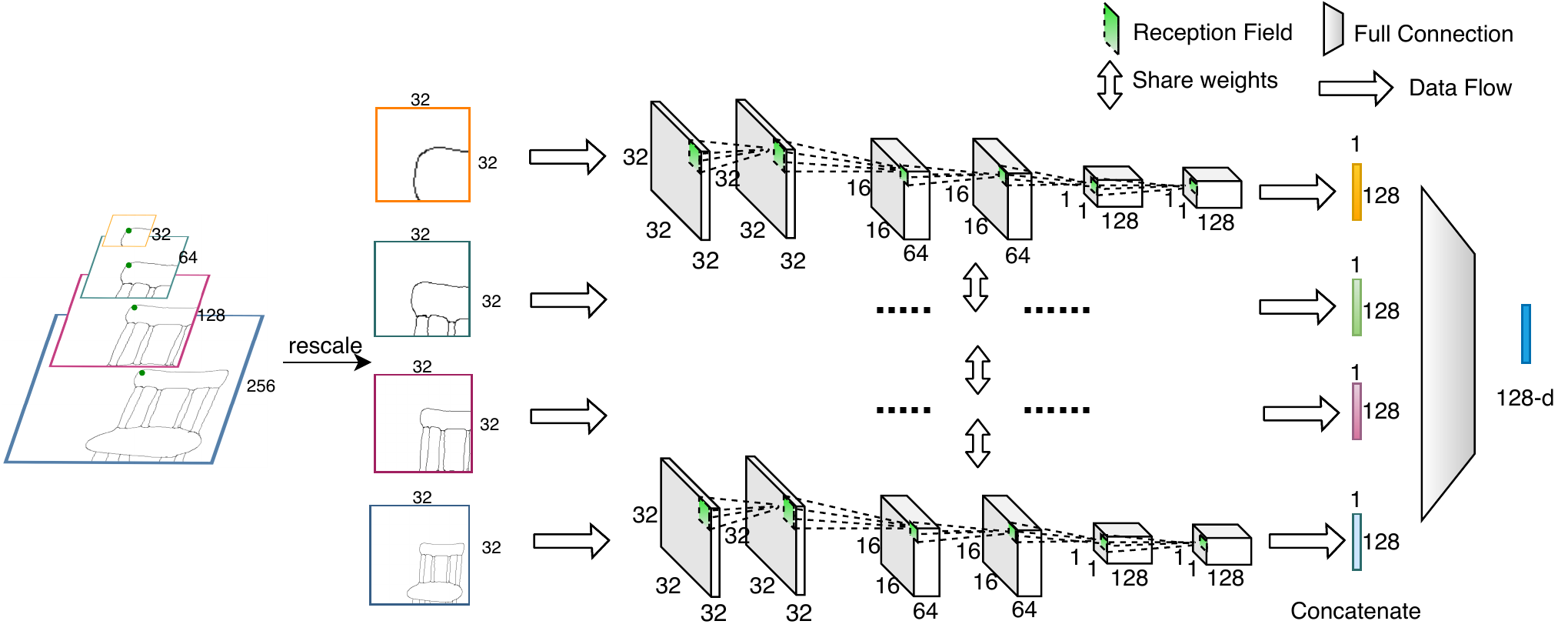}
\end{center}
\caption{The architecture of \sysName-Net. Our input is a four-scale patch pyramid (32$\times$32, 64$\times$64, 128$\times$128, 256$\times$256) centered at a pixel of interest on a sketch, with each scale 
rescaled to 32$\times$32. Given the multi-scale patches, we design a multi-branch framework with shared weights to take as input these rescaled patches. The dashed lines represent the data flow from an input patch to an output descriptor.
For the kernel size and stride in our network, we adopt the same settings as \cite{tian2017l2}. Finally, the output as a 128-D descriptor embeds features from the four scales by concatenation and full connection operations.}
\label{fig:Figure Network}
\end{figure*}

\textit{Terminology.} 
In the next sections, we differentiate between the word ``points'' for shapes, sketch images, and sketch lines as follows. We refer to the points on the surface of a 3D shape as \emph{vertices}; the points on a 2D sketch image as \emph{pixels} and the points exactly on the sketch lines as \emph{points}.

For input sketches represented as rasterized images, a key problem to semantic correspondence learning is to measure the difference between a pair of pixels in a semantic way. 
This is essentially a metric function learning problem where a pair of corresponding pixels has a smaller distance than a pair of non-corresponding {pixels} in the metric space. Formally, it can be expressed as follows: 
\begin{equation}
   \label{eqn:01}
   D_m\langle{p_{c1}},{p_{c2}}\rangle < D_m\langle{p_{c1}},{\widetilde{p}_{nc}}\rangle, 
\end{equation}
where $D_m\langle{{\cdot}, {\cdot}}\rangle$ is a metric function, and $p_{c1}$ and $p_{c2}$ represent corresponding pixels from different sketches (e.g., the pair of anchor and positive pixels in Figure \ref{fig:Figure TripLets}) while $p_{c1}$ and $\widetilde{p}_{nc}$ are a pair of non-corresponding {pixels} (e.g., the pair of anchor and negative pixels in Figure \ref{fig:Figure TripLets}).

We follow \cite{tian2017l2}
to build the metric function $D_m$ by learning a sketch descriptor (\sysName) with a triplet loss function. 
Basically we optimize the loss function in Equation \ref{eqn:02}, which minimizes the distance between pairs of corresponding pixels and maximizes the distance between pairs of non-corresponding pixels.
Since 
sketch images are rather sparse, we adopt a multi-scale patch-based representation, that is, multi-scale patches centered at a pixel of interest in a sketch.
We resort to deep CNNs, which own the superior capability of learning discriminative visual features from sufficient training data. 
As illustrated in Figure \ref{fig:Figure Network}, our designed network (Section \ref{subsection:Network Architecture}) takes a multi-scale patch as input and outputs a 128-D distinctive descriptor.

To train the network, we first synthesize line drawings of a 3D shape from different viewpoints as multi-view sketches.
We then generate the ground-truth correspondences by first uniformly sampling points on the 3D shape and then projecting them to the corresponding multiple views. We will discuss the process of data preparation in more detail
in the next section.

\subsection{Data Preparation}
\label{subsection:Data Preparation}

\textit{Multi-view Sketches with Ground-truth Correspondences.}
\label{subsubsection:Preserve 3D Corresponding Relationships} 
We follow a similar strategy in \cite{li2018fast,xu2013sketch2scene,su2015multi} to synthesize sketches from 3D shapes. Specifically, we first render a 
3D shape with the aligned upright orientation to a normal map {under a specific viewpoint}, and then extract an edge map from the normal map using Canny edge detection. We adopt this approach instead of the commonly used suggestive contours \cite{decarlo2003suggestive}, because the latter is suited for high-quality 3D meshes and cannot generate satisfactory contours from 
poorly-triangulated meshes (e.g., airplanes and rifles in the Structure Recovery dataset \cite{shen2012structure}). Hidden lines of the edge detection results are removed. In our implementation, each sketch is resized to a $480 \times 480$ image.
As mentioned in \cite{eitz2012humans}, most humans are not faithful artists and create sketches in a casual and random way.
Unlike \cite{huang2018learning,navarro2019sketchzooms} using limited views, to better accommodate the shape and viewpoint variations in freehand sketches,
we sample viewpoints on the upper unit viewing hemisphere (in {an} elevation angle of 15$\sim$45 degrees) at every 15 or 30 degrees in the azimuth angle for rendering each 3D model.

By projecting each vertex to the corresponding views, we naturally construct ground-truth correspondences (with the projections from the same vertices) among synthesized multi-view sketches, as illustrated in Figure \ref{fig:Figure dataPineline}. We do not consider hidden vertex projections (invisible under depth testing).
If the projections of a vertex are visible only in less than two different views, this vertex is not considered in ground-truth correspondences.
Following this strategy, we can generate from 28K to 60K ground-truth pairwise correspondences from each 3D shape.
Our synthesized correspondence dataset for training and testing are derived from 6,852 multi-view sketches distributed over 18 shape categories of existing shape datasets~\cite{shen2012structure,chen2009benchmark,yi2016scalable} (Section~\ref{section:Experiments}).  
We will make our dataset publicly available.

\begin{figure}[t]
   \begin{center}
   \includegraphics[width=0.8\linewidth]{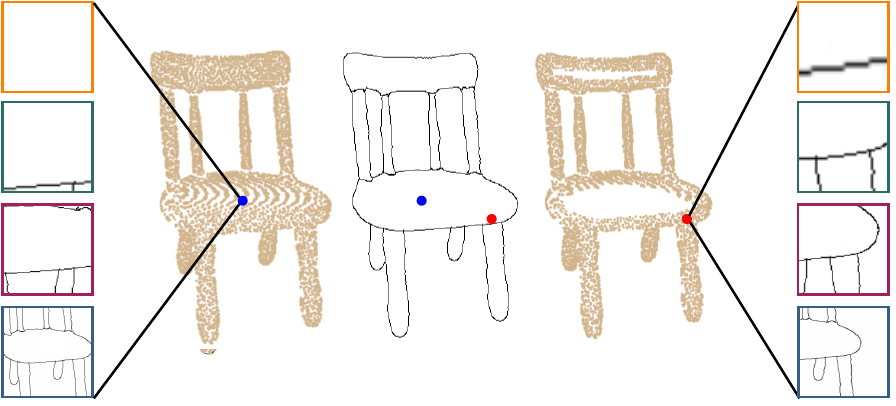}
   \end{center}
      \caption{
      An illustration of two sampling mechanisms (Left: OR-sampling; Right: AND-sampling) for training data.
      OR-sampling generates more challenging data (blue) than AND sampling (red).
      }
   \label{fig:Figure FilterBinaryMap}
   \end{figure}

\textit{Multi-scale Patch Representation.}
We represent each visible projection of a 3D vertex on sketch images with a patch-based representation centered at the corresponding pixel to capture the distinctive neighboring structures (Figure \ref{fig:Figure TripLets}).
To better handle the sparsity and the lack of texture information in sketches, we adopt a multi-scale strategy. Given a $480 \times 480$ sketch image, we employ a four-scale representation (i.e., $32 \times 32$, $64 \times 64$, $128 \times 128$, and $256 \times 256$) for a pixel, as illustrated in Figure \ref{fig:Figure TripLets}.

The multi-scale patch-based representation allows us to sample ground-truth correspondences inside sketched objects, and not necessarily on sketch lines.
This significantly increases the number of ground-truth correspondences for training.
However, we do require valid information existing in a multi-scale patch representation. 
We have tried two sampling mechanisms: 1) \emph{OR-sampling}: a multi-scale patch is valid if the patch is non-empty at any of the scales (Figure \ref{fig:Figure FilterBinaryMap} left); 2) \emph{AND-sampling}: a multi-scale is valid if the patch is non-empty at every scale (Figure \ref{fig:Figure FilterBinaryMap} right). 
The former generates valid multi-scale patches at almost every visible vertex projection,
since the patch at scale $256 \times 256$ is often non-empty given its relatively large scale. 
In contrast, the AND-sampling generates valid multi-scale patches only near to sketch lines.
We will compare the performance of these two sampling mechanisms in Section \ref{section:Experiments}.

Before feeding these patches into our multi-branch network, we rescale all the patches to 32 $\times$ 32 (i.e., the smallest scale) by {bilinear interpolation (Figure \ref{fig:Figure Network}). Below we describe our network architecture in more detail.

\subsection{Network Architecture}
\label{subsection:Network Architecture}
We design a network architecture to learn local descriptors for measuring the semantic distance between a pair of pixels
in multi-view sketches.
As illustrated in Figure \ref{fig:Figure Network}, our network has four branches to process the set of four scaled input patches.
The four branches share the {same architecture and weights} in the whole learning process. 
Each branch receives a 32 $\times$ 32 patch and outputs a 128 $\times$ 1 (i.e., 128-D) feature vector which is then further fused at the concatenation layer and the final fully-connected layer. Note that due to the shared weights among the branches, the multi-branch structure does not increase the number of parameters.
Our network produces a 128-D descriptor as output, which will be later used for sketch correspondence and pixel-wise retrieval  
in Section \ref{section:Experiments}.

\subsection{Objective Function}
\label{subsection:Objective Function}
To train our network, we employ the random sampling strategy of \cite{tian2017l2} to assemble the triplets (Figure \ref{fig:Figure TripLets}) in a training batch.
We define the output descriptors of a triplet as ($f_{a}, f_{p}, f_{n}$), where $f_{a}$ is the descriptor vector of an anchor pixel in one sketch, and $f_{p}$ and $f_{n}$ represent the descriptors of the corresponding pixel (corresponding to the same vertex in a 3D shape as the anchor pixel) and a non-corresponding pixel. The non-corresponding pixel can be selected from either the same sketch or in the other views.
With the descriptor triplets, we adopt the triplet loss \cite{schroff2015facenet} to train the network. The triplet loss, given in Equation \ref{eqn:02}, aims to pull closer the distance between a pair of corresponding {pixels} ($f_{a}, f_{p}$) and push away the distance between a pair of non-corresponding {pixels} ($f_{a}, f_{n}$) in the metric space.
\begin{equation}
\label{eqn:02}
L_{triplet}= {\frac{1}{n}}\Sigma_{i=1}^{n} \max\left(0,  d\left(f_{a_i}, f_{p_i}\right)  -  d\left(f_{a_i}, f_{n_i}\right)  +  m \right),
\end{equation}
where $n$ is the number  
of triplets in a training batch, $\left(f_{a_i}, f_{p_i}, f_{n_i}\right)$ denotes the $i$-th triplet,  $d\left({\cdot}, {\cdot}\right)$ measures the Euclidean distance given two descriptors,
 and the margin $m$ is set to $1.0$ in our experiments. 

   \begin{figure}[t]
      \begin{center}
      \includegraphics[width=1\linewidth]{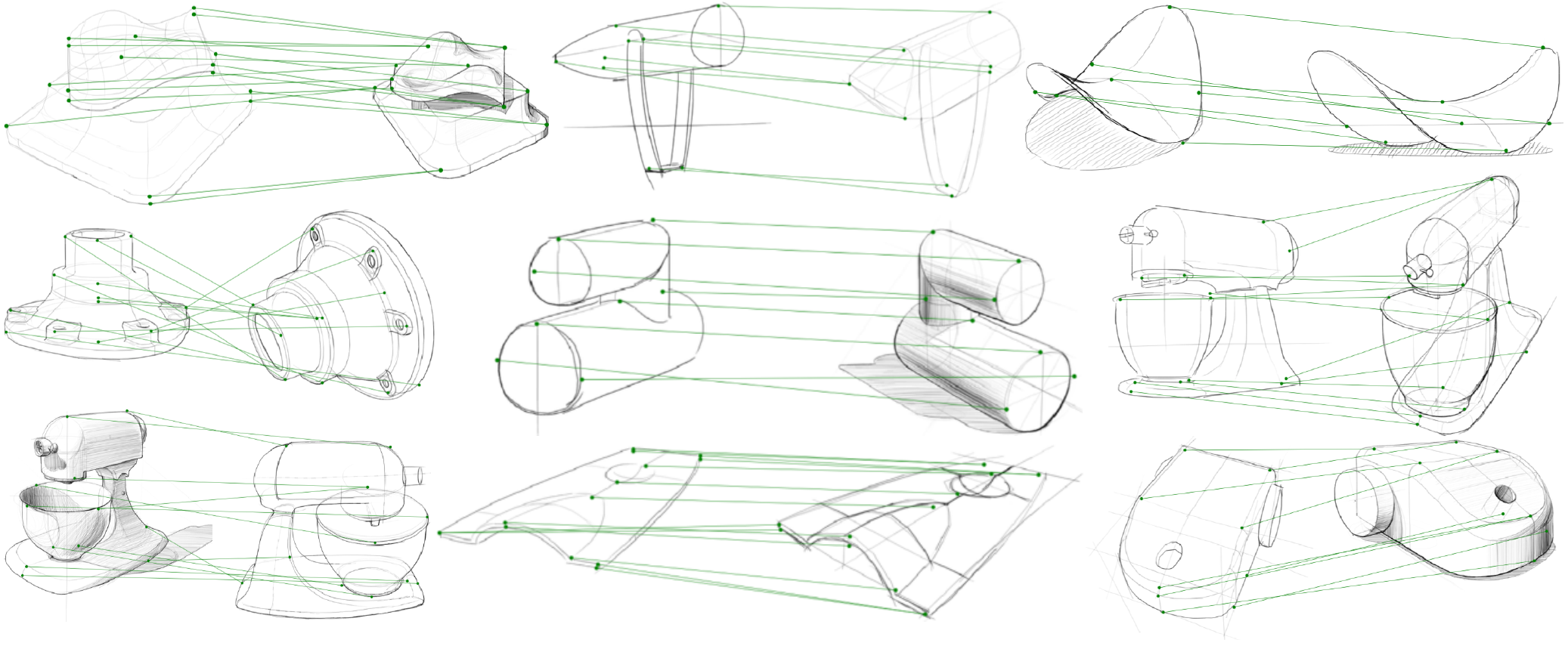}
      \end{center}
         \caption{Sketch correspondence in the OpenSketch dataset computed with \sysName~descriptors. Note that even with limited ground-truth correspondence in the training set of OpenSketch, \sysName~can still establish a robust correspondence for the 
         multi-view sketches.
         }
      \label{fig:Figure OpenSketchCorrespondence}
   \end{figure}   
\begin{table*}
   \begin{center}
    \resizebox{0.9\textwidth}{!}{
   \begin{tabular}{|l|c|c|c|c|c|c|c|c|c|c|c|c|c|c|c|c|c|c|c|c|c|c|c|c|}
   \hline
   & \multicolumn{6}{|c|}{Structure-Recovery} & \multicolumn{9}{|c|}{PSB} & \multicolumn{8}{|c|}{ShapeNet}\\
   \hline
   {Category} &\rotatebox{90}{Airplane}&\rotatebox{90}{Bycicle}& \rotatebox{90}{Chair}&\rotatebox{90}{Fourleg}&\rotatebox{90}{Human}&\rotatebox{90}{Rifle}&\rotatebox{90}{Airplane}&\rotatebox{90}{Bust}&\rotatebox{90}{Chair}&\rotatebox{90}{Cup}&\rotatebox{90}{Fish}&\rotatebox{90}{Hand}&\rotatebox{90}{Human}&\rotatebox{90}{Octopus}&\rotatebox{90}{Plier}&\rotatebox{90}{Airplane}&\rotatebox{90}{Bag}&\rotatebox{90}{Cap}&\rotatebox{90}{Car}&\rotatebox{90}{Chair}&\rotatebox{90}{Earphone}&\rotatebox{90}{Mug}&\rotatebox{90}{Pistol}\\ 
   \hline
   Shapes &54&12&27&16&16&25&20&20&20&20&20&20&20&20&20&38&35&35&39&23&28&34&25\\ 
   \hline
   Views &11&12&12&12&12&12&11&12&12&11&12&11&12&11&11&11&12&12&12&12&12&12&12\\ 
   \hline
   \end{tabular}
   }
   \end{center}
   \caption{
   Object categories and the number of shapes and views used in our dataset.
   }
   \label{table:tabCateAndNum}
   \end{table*}

\section{Experiments}
\label{section:Experiments}
 
We conducted extensive experiments on two sketch datasets: our synthesized multi-view sketch dataset and the OpenSketch dataset\cite{gryaditskaya2019opensketch}.
For our synthesized dataset {(Figure \ref{fig:Figure testVsTrain})}, we utilize three existing 3D shape repositories: the Structure-Recovery database \cite{shen2012structure}, Princeton Segmentation Benchmark (PSB) \cite{chen2009benchmark}, and ShapeNet \cite{yi2016scalable}.
Table \ref{table:tabCateAndNum} shows some information about the three shape repositories, including the selected categories, the number of 3D shapes per category, and  the number of  views used per category.
The shapes in each category were manually selected based on the criterion of increasing shape diversity and decreasing the redundant and repeated shapes.
For the OpenSketch dataset, it has around 400 sketches of 12 man-made objects, which were drawn by professional product designers.
Different from the synthesized sketches, the OpenSketch data contains abundant annotations of additional shadings, skeletons and auxiliary lines (shown in Figure \ref{fig:Figure matchingReults} (bottom) and Figure \ref{fig:Figure OpenSketchCorrespondence}).

In addition to the hand-drawn sketches in OpenSketch,
we also performed evaluation on a set of
more freely-drawn sketches.
Several participants were invited to create freehand sketches on a touchscreen,
after observing a 3D shape for a fixed amount of time. Each participant was given three salient views of each shape that ordinary users would be familiar with. Figure \ref{fig:Figure matchingReults} (two middle rows) shows some representative results of sketch correspondence (i.e., chair, bicycle, and fourleg). Note that to avoid visual clutter,
for each freehand
sketch pair, we randomly selected 20$\sim$50 pairs of matched correspondences computed by nearest neighbor search with \sysName
 ~(see Section~\ref{subsubsection:Sketch Correspondence}).
In the supplementary material, we provide more results of the computed correspondences among hand-drawn multi-view sketches.

\subsection{Implementation Detail}
\label{subsection:Implementation Details}

We implemented our network with the PyTorch \cite{paszke2019pytorch} framework and used the Xavier initialization \cite{glorot2010understanding}.
We train our network {for each object category} with a data splitting ratio of 8 : 1 : 1 (training : validation : testing).
All multi-view sketches are rendered to the size of $480\times480$. 
The batch size is set to $64$.
Our network is trained on an NVIDIA RTX $2080$Ti GPU and optimized by the Adam {\cite{kingma2014adam}} optimizer ($\beta_1 = 0.9$ and $\beta_2 = 0.999$) with a learning rate of $1e^{-3}$. The number of iteration epochs in our experiments is set to $100$.

\subsection{Performance Evaluation}
\label{subsection:Performance Metrics}

To verify the effectiveness of our proposed network, 
we design two evaluation tasks: multi-view sketch correspondence and pixel-wise
retrieval. We compare our approach with the existing learning-based descriptors, 
including
LeNet \cite{lecun1998gradient}, L2-Net \cite{tian2017l2}, HardNet \cite{mishchuk2017working}, SOSNet \cite{tian2019sosnet} and AlexNet-based view pooling \cite{huang2018learning, navarro2019sketchzooms} (AlexNet-VP in short). Note that we reimplemented LeNet, L2-Net, HardNet, SOSNet, and AlexNet-VP following their original configurations.
For AlexNet-VP, we use the same multi-scale patch inputs as our network.
The training input to all other networks is a single-scale 32$\times$32 patch. For a fair comparison, we only consider pixels that pass the AND-sampling criteria, that is, the 32 $\times$ 32 patch of each multi-scale input should be non-empty.

\begin{figure}[t]
   \begin{center}
   \includegraphics[width=1\linewidth]{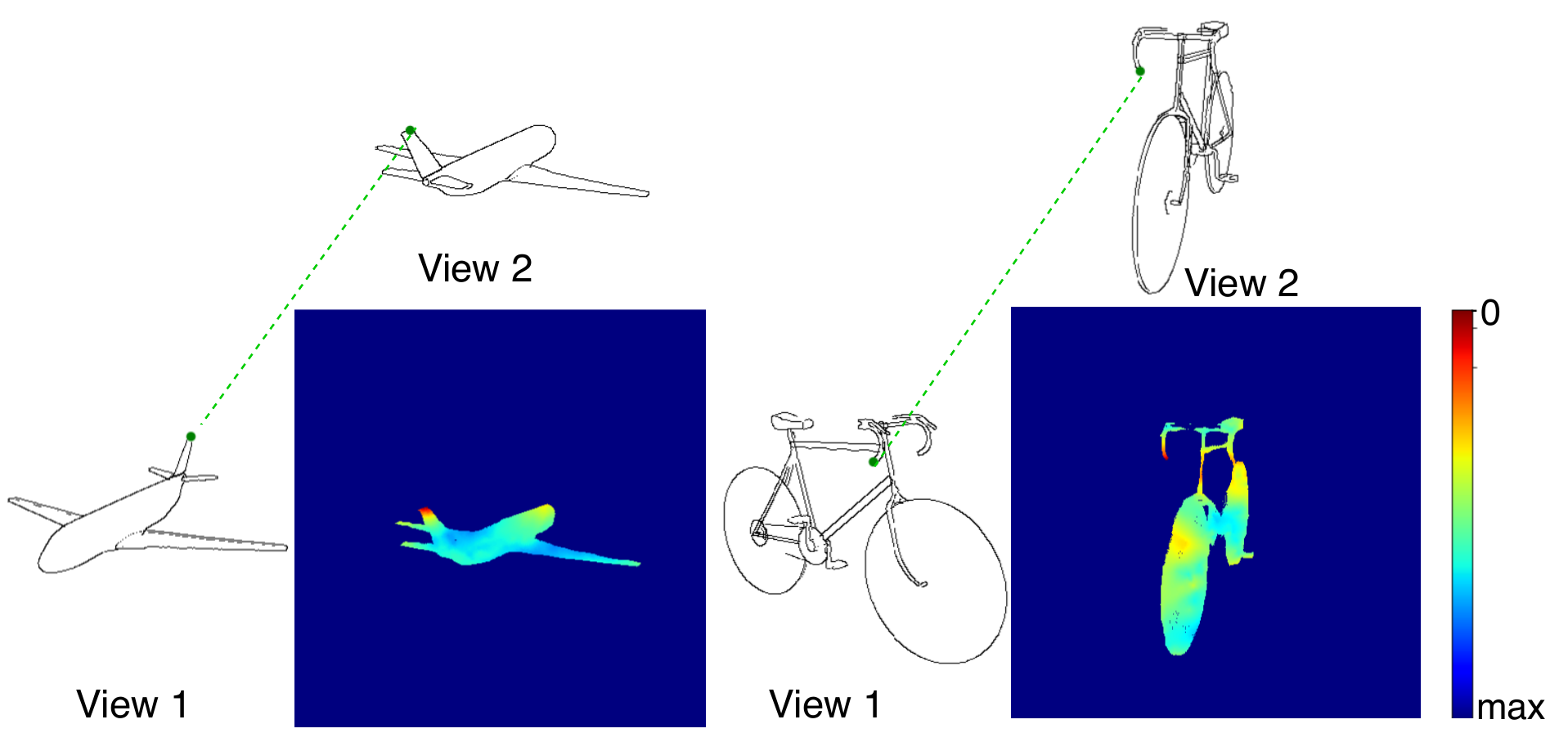}
   \end{center}
      \caption{
      For a given pixel inside a sketched object under View 1,
      we find a corresponding point in the other sketch under View 2 by computing a distance map through our learned descriptor.  
      }
   \label{fig:Figure DistanceVisualization}
   \end{figure}
\begin{figure}[t]
   \begin{center}
   \includegraphics[width=1\linewidth]{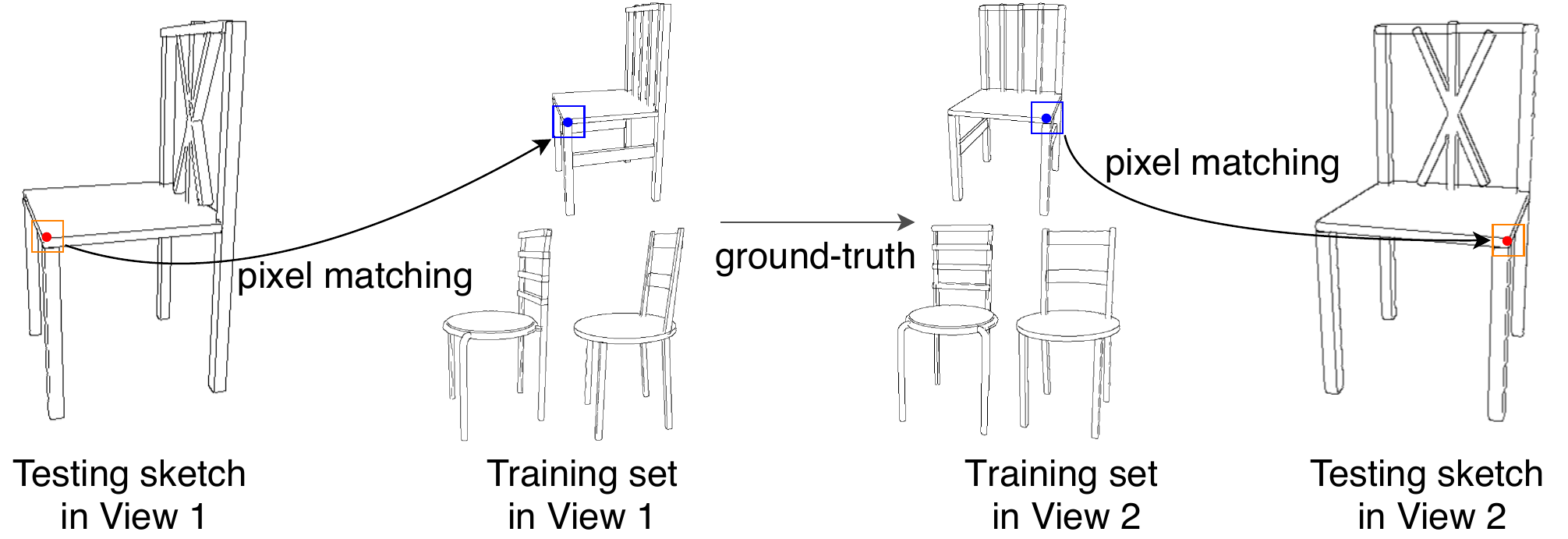}
   \end{center}
      \caption{
            The pipeline of a retrieval-based baseline, which directly utilizes
            the ground-truth correspondence in the training set. For illustration, we show the top-3 sketches retrieved from the training set by pixel matching.
      }
   \label{fig:Figure baseline}
   \end{figure}
\begin{figure*}
   \begin{center}
   \includegraphics[width=1\linewidth]{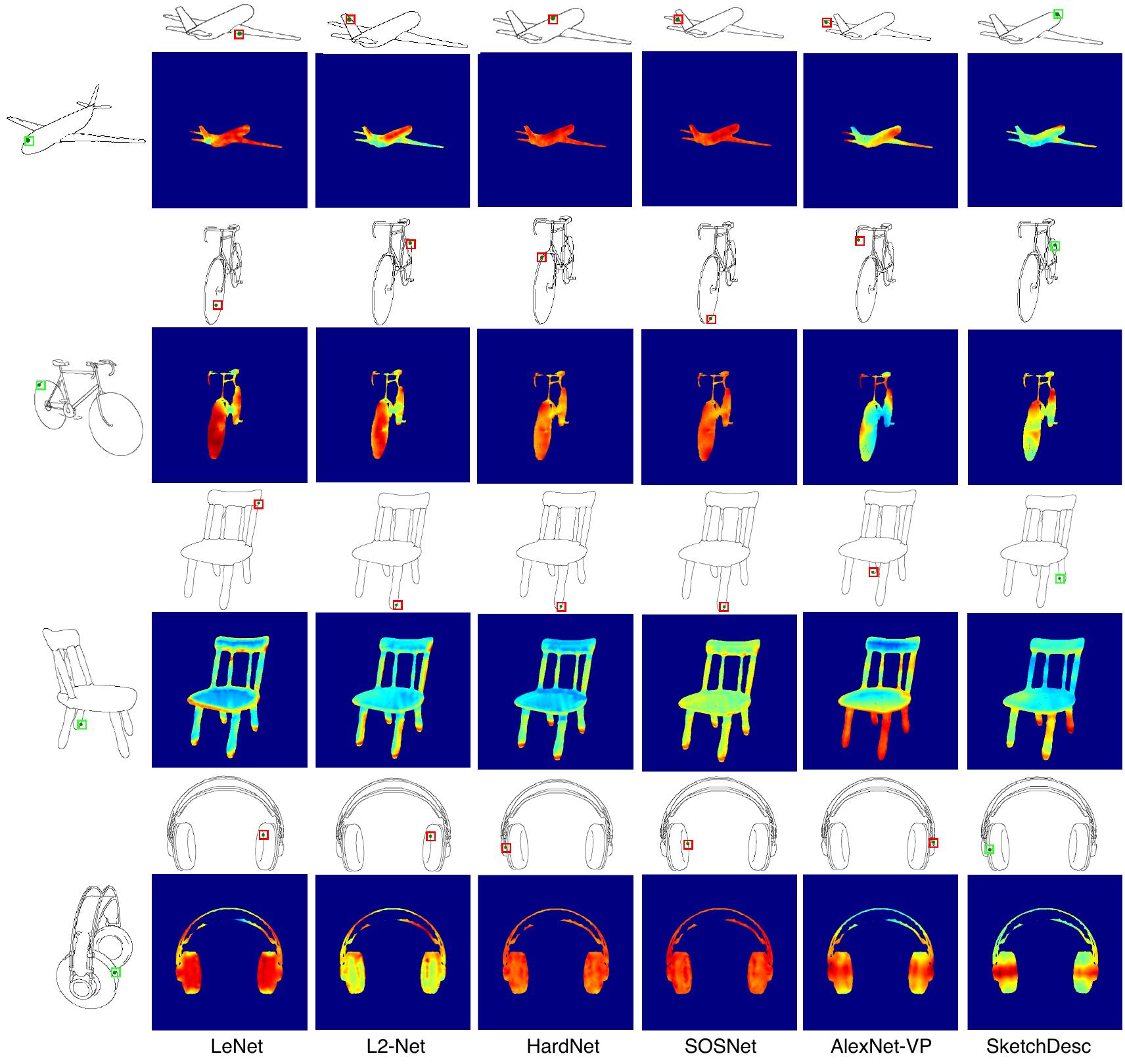}
   \end{center}
      \caption{
      Sketch correspondence results by computing the distance maps with different approaches. Correct and wrong matching results are marked as green and red boxes, respectively. 
}
   \label{fig:Figure distanceComparison}
   \end{figure*}
   
\subsubsection{Sketch Correspondence}
\label{subsubsection:Sketch Correspondence}
In this task, we validate the performance of the learned descriptors in finding corresponding pixels in pairs of multi-view sketches in the testing set.
Given a pair of testing sketches, for each pixel 
in one sketch, we compute its distances to all pixels in the other sketch (see distance visualization in Figure \ref{fig:Figure DistanceVisualization}).
We consider it as a successful matching if the pixel with the shortest distance is no further than 16 pixels (half of the smallest patch scale) away from the ground-truth pixel. Note that not all pixels in a sketch image have ground-truth correspondences (only among those projected).

To clarify the dissimilarity between the training set and the testing set, we  designed a retrieval-based baseline (illustrated in Figure \ref{fig:Figure baseline}) in the task of sketch correspondence.

For a given pixel on one sketch in View 1 of a testing pair, we perform the pixel matching operation (by nearest neighbor search) to get the most similar pixel among all the training sketches under View 1. 
Note that the pixel matching operation takes as input the patches (32$\times$32) centered on the pixels.
With the matched pixel, we utilize the ground-truth cross-view correspondence in the training data to find its corresponding pixel in View 2 (i.e., the same view as the other sketch in the testing sketch pair).
Finally, we perform the pixel matching again to find the most similar pixel on the testing sketch in View 2.
Note that we employ the conventional image matching method \cite{dalal2005histograms} for the pixel matching operation.

The performance of our approach and other competing methods on several representative categories is reported in Table \ref{table:tabshortResultsAlignments}.
We report the averaged success rate as the correspondence accuracy.
We observe that our network outperforms its learning-based and retrieval-based competitors.
Due to the lack of texture in sketch patches, the frameworks designed for image patches 
(i.e., L2-Net, HardNet, and SOSNet) cannot learn effective descriptors to match the corresponding pixels in multi-view sketches. Our descriptor also surpasses those based on LeNet and AlexNet-VP. Figure \ref{fig:Figure distanceComparison} gives some qualitative comparisons between different approaches. Since 32 $\times $ 32 patches are required to be non-empty, the testing pixels are near sketch lines.
We observe that the descriptors learned by LeNet, HardNet, L2-Net, and SOSNet are less discriminative, and are thus not effective in finding corresponding pixels among multi-view sketches.
AlexNet-VP uses the max-pooling layer, making it difficult to focus on local details for robust descriptors, as discussed in \cite{mishchuk2017working}.
As a consequence, AlexNet-VP produces more ambiguous distance maps (e.g., the chair example in Figure \ref{fig:Figure distanceComparison}) compared to our method.
The retrieval-based baseline employs the ground-truth from the training set, but it still fails to build reliable correspondences for multi-view sketches. 
{In fact,} the retrieval-based baseline has the worst performance.
It is mainly because of
the incapability of the image-based descriptor~\cite{dalal2005histograms} on dealing with the highly similar local patches with limited textures in multi-view sketches,
and also because of
the large disparity between multi-view sketches in the training set and testing set (Figure \ref{fig:Figure testVsTrain}).
In addition, it is challenging to use this baseline in practice with hand-drawn 
sketches, due to the required view prior and its large running time to infer the correspondences for a pair of multi-view sketches (3.15 minutes vs 8.76 seconds by our method on average).
\begin{table}[t]
   \begin{center}
      \resizebox{0.48\textwidth}{!}{
   \begin{tabular}{|c|c|c|c|c|c|c|c|c|c|}
   \hline
   &&base.&LeNet&L2\-Net&HardNet&SOSNet&AlexNet-VP&SketchDesc\\
   \hline
   \multirow{7}{*}{\tabincell{c}{Structure-\\Recovery}} 
   &Airplane&{0.14}&0.18&0.24&0.21&0.19&0.36&\textbf{0.54}\\
   &Bicycle&{0.28}&0.38&0.39&0.42&0.40&0.66&\textbf{0.71}\\
   &Chair&{0.27}&0.40&0.40&0.39&0.39&0.51&\textbf{0.56}\\
   &Fourleg&{0.27}&0.29&0.34&0.32&0.28&0.52&\textbf{0.66}\\
   &Human&0.21&0.28&0.38&0.40&0.35&0.64&\textbf{0.73}\\
   &Rifle&0.27&0.43&0.52&0.53&0.50&0.67&\textbf{0.76}\\
   \cline{2-9}
   &\textbf{avg\_acc}&{0.24}&0.33&0.38&0.38&0.35&0.56&\textbf{0.66}\\
   \hline
   
   \multirow{10}{*}{PSB}
   &Airplane&{0.07}&0.18&0.23&0.13&0.10&0.52&\textbf{0.72}\\
   &Bust&0.20&0.33&0.22&0.23&0.20&0.31&\textbf{0.45}\\
   &Chair&0.22&0.29&0.27&0.24&0.26&0.44&\textbf{0.60}\\
   &Cup&0.26&0.22&0.25&0.19&0.22&0.35&\textbf{0.57}\\
   &Fish&{0.29}&0.30&0.36&0.35&0.31&0.63&\textbf{0.64}\\
   &Human&0.20&0.35&0.35&0.38&0.34&0.54&\textbf{0.79}\\
   &Octopus&{0.08}&0.10&0.11&0.09&0.07&0.24&\textbf{0.28}\\
   &Plier&0.09&0.10&0.09&0.06&0.14&0.57&\textbf{0.73}\\
   \cline{2-9}
   &\textbf{avg\_acc}&0.18&0.23&0.24&0.21&0.21&0.45&\textbf{0.62}\\
   \hline

   \multirow{8}{*}{ShapeNet}
   &Airplane&{0.21}&0.39&0.44&0.40&0.35&0.67&\textbf{0.69}\\
   &Bag&{0.11}&0.17&0.14&0.15&0.14&0.31&\textbf{0.38}\\
   &Cap&0.17&0.21&0.25&0.26&0.22&0.55&\textbf{0.75}\\
   &Car&0.13&0.17&0.21&0.21&0.21&0.55&\textbf{0.73}\\
   &Chair&{0.16}&0.19&0.20&0.18&0.19&0.42&\textbf{0.47}\\
   &Earphone&0.12&0.15&0.18&0.16&0.14&0.25&\textbf{0.42}\\
   &Mug&0.21&0.19&0.20&0.17&0.17&0.32&\textbf{0.39}\\
   &Pistol&{0.23}&0.30&0.34&0.33&0.34&0.66&\textbf{0.71}\\
   \cline{2-9}
   &\textbf{avg\_acc}&{0.17}&0.22&0.25&0.23&0.22&0.47&\textbf{0.57}\\
   \hline
   \end{tabular}
}
    \end{center}
    \caption{Sketch correspondence accuracy (i.e., the average success rate) for the different methods. The best results in each object category are in boldface.
  }
  \label{table:tabshortResultsAlignments}
\end{table}

\begin{table}
   \label{table:appendtabretrievaldata}
   \begin{center}
   \resizebox{0.48\textwidth}{!}{
   \begin{tabular}{|c|c|c|c|c|c|c|c|c|}
   \hline
   &&LeNet&L2-Net&HardNet&SOSNet&AlexNet-VP&SketchDesc\\
   \hline
   \multirow{7}{*}{\tabincell{c}{Structure-\\Recovery}} 
   &Airplane&0.33&0.42&0.45&0.40&0.45&\textbf{0.68}\\
   &Bicycle&0.37&0.40&0.44&0.39&0.60&\textbf{0.84}\\
   &Chair&0.39&0.45&0.42&0.42&0.70&\textbf{0.82}\\
   &Fourleg&0.33&0.45&0.41&0.30&0.75&\textbf{0.82}\\
   &Human&0.20&0.44&0.40&0.27&0.74&\textbf{0.92}\\
   &Rifle&0.56&0.56&0.57&0.54&0.76&\textbf{0.83}\\
   \cline{2-8}
   &\textbf{avg\_map}&0.36&0.45&0.45&0.39&0.67&\textbf{0.82}\\
   \hline
   \multirow{10}{*}{PSB}
   &Airplane&0.26&0.26&0.11&0.11&0.42&\textbf{0.68}\\
   &Bust&0.01&0.29&0.29&0.26&0.51&\textbf{0.64}\\
   &Chair&0.33&0.48&0.42&0.46&0.75&\textbf{0.86}\\
   &Cup&0.05&0.08&0.04&0.06&0.22&\textbf{0.50}\\
   &Fish&0.19&0.28&0.23&0.21&0.58&\textbf{0.73}\\
   &Human&0.13&0.53&0.52&0.46&0.79&\textbf{0.93}\\
   &Hand &0.08&0.37&0.37&0.28&0.52&\textbf{0.64}\\
   &Octopus&0.38&0.29&0.29&0.46&0.67&\textbf{0.71}\\
   &Plier&0.18&0.36&0.21&0.18&0.79&\textbf{0.86}\\
   \cline{2-8}
   &\textbf{avg\_map}&0.18&0.33&0.28&0.27&0.58&\textbf{0.73}\\
   \hline
   \multirow{8}{*}{ShapeNet}
   &Airplane&0.14&0.22&0.15&0.09&0.37&\textbf{0.55}\\
   &Bag&0.17&0.20&0.15&0.14&0.53&\textbf{0.69}\\
   &Cap&0.16&0.24&0.19&0.17&0.59&\textbf{0.71}\\
   &Car&0.20&0.19&0.22&0.22&0.56&\textbf{0.67}\\
   &Chair&0.11&0.15&0.14&0.140&0.51&\textbf{0.67}\\
   &Earphone&0.25&0.30&0.25&0.20&0.49&\textbf{0.83}\\
   &Mug&0.05&0.06&0.06&0.08&0.28&\textbf{0.34}\\
   &Pistol&0.29&0.35&0.35&0.36&0.69&\textbf{0.82}\\
   \cline{2-8}
   &\textbf{avg\_map}&0.17&0.21&0.19&0.17&0.50&\textbf{0.66}\\
   \hline
   \end{tabular}
   }
\end{center}
 \caption{
   Pixel-wise retrieval performance for the different methods.}
   \label{table:tabshortRetirvalData}
\end{table}
 \begin{figure}[t]
   \begin{center}
   \includegraphics[width=1\linewidth]{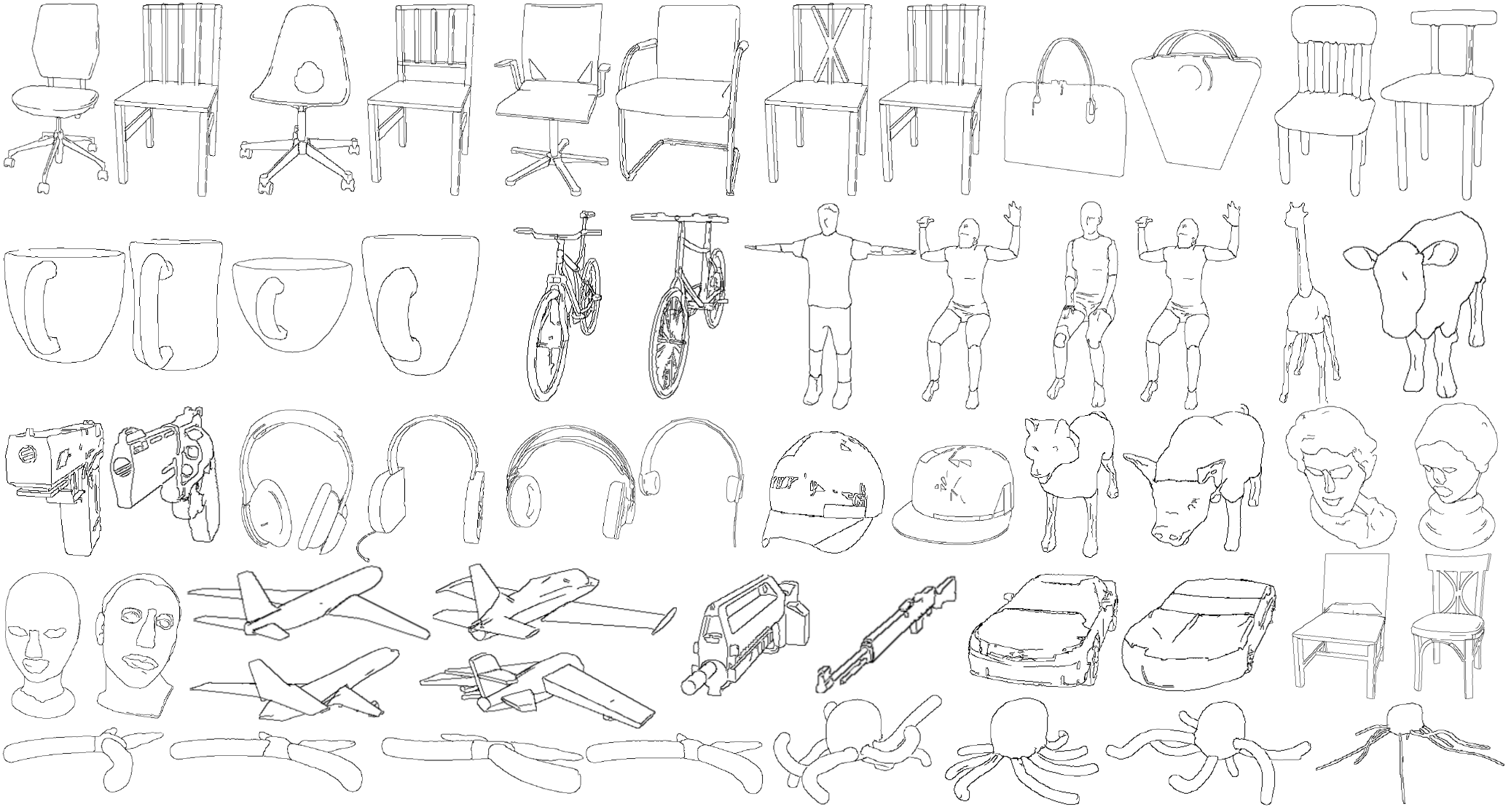}
   \end{center}
    \caption{The pairs of a testing sketch (left) and its most similar training sketch (right) retrieved from the training set under the same view, utilizing the image matching method \cite{dalal2005histograms}.
      }
   \label{fig:Figure testVsTrain}
   \end{figure}
Figure \ref{fig:Figure matchingReults} shows successful matchings randomly selected from all successful matchings 
on the synthesized multi-view sketches (top) and the hand-drawn sketches (middle).
In Figures \ref{fig:Figure matchingReults}(bottom) and \ref{fig:Figure OpenSketchCorrespondence}, we also show that
\sysName~can be further utilized to infer some correspondences of multi-view sketches in the OpenSektch dataset. Due to the limited ground-truth correspondence in OpenSketch, we show all the successful matchings.
More correspondence results (successful and unsuccessful matchings) of multi-view sketches can be found in the supplementary material.

\subsubsection{Multi-view Pixel-wise Retrieval}
\label{subsubsection:Corresponding Point Retrieval}
We further design a multi-view corresponding pixel retrieval task. Given multi-view sketches synthesized from multiple shapes, we uniformly sample a set of pixels (1000$\sim$1600 pixels) on one sketch and search in the other sketches for the corresponding pixels which are from the same shape (in different views).
We use the descriptors computed from the compared networks as queries and adopt the Mean Average Precision (MAP)~\cite{hpatches_2017_cvpr} to measure the retrieval performance.
Let $\bm{y} = (y_1,\dots,y_n) \in \{-1,+1\}^n$ be the labels of a ranked list of pixels returned for a pixel query, with -1 and +1 indicating negative and positive match,  respectively. Then the \emph{precision} at rank $i$ is given by\footnote{Here $[z]_+ = \max\{0,z\}$.}
$
P_i(\bm{y}) =
  {\sum_{k=1}^i [y_k]_+}/{\sum_{k=1}^i |y_k|}
$ {and} 
the \emph{average precision} (AP) is given by 
$
  AP(\bm{y}) = \sum_{k:y_k=+1} P_k(\bm{y}) /\sum_{k=1}^N [y_k]_+. 
$
Finally, given $N_q$ as the number of total query pixels, the mean average precision (MAP) is computed by
$
MAP =\sum_{y=1}^{N_q} AP_y / N_q
$.
Experiments are performed category-wise in Structure-Recovery, PSB, and ShapeNet. 
To further report the performance on a dataset, we adopt $avg\_map$ which averages the MAP over all the tested categories.

Table \ref{table:tabshortRetirvalData} shows the results given by the compared methods. 
\sysName~achieves the best performance among all the learned descriptors.
Our learned descriptor surpasses the image-based descriptors of L2-Net, HardNet, and SOSNet by a large margin. For AlexNet-VP, it achieves a closer but still lower performance compared with our method.

\subsubsection{Cross-dataset Validation}
Considering the limited data in the testing set, we further demonstrate the generalization ability of \sysName~with a cross-dataset validation.

For the three shape datasets (Structure-Recovery, PSB, and ShapeNet), we take two of them as the training set and the remaining one as the testing set.
To reduce the effects of imbalanced data distribution, we report the performance on a more balanced and overlapping category (Chair), which has 27, 20, and 23 shapes in each of the three datasets respectively (Table \ref{table:tabCateAndNum}).
We report the performance of the sketch correspondence task and the multi-view pixel-wise retrieval task in Table \ref{table:tabcrossdatasetcross} and Table \ref{table:tabcrossdatasetretrival} respectively.
We observe that \sysName~still shows an overwhelming superiority over its competitors.
There is only a slight drop (0.02 on average)
in the cross-dataset performance of the sketch correspondence task, compared to Table \ref{table:tabshortResultsAlignments}.
However, the cross-dataset performance of the pixel-wise retrieval task shows a noticeable degradation (0.15 on average), compared to Table \ref{table:tabshortRetirvalData}.
This is mainly because that the pixel-wise retrieval task
has a much larger search space (multi-view sketches of multiple shapes) than the sketch correspondence task (paired multi-view sketches of a single shape), and
tends to lead to more mismatches due to ambiguity.
In general, the robustness and effectiveness of \sysName~are further verified by the large and unseen testing data.
\begin{table}
   \begin{center}
       \resizebox{0.505\textwidth}{!}{
    \begin{tabular}{|l|c|c|c|}
    \hline
    Methods&\tabincell{c}{\underline{Structure-Recovery\&ShapeNet}\\PSB} & \tabincell{c}{\underline{PSB\&ShapeNet}\\Structure-Recovery} &\tabincell{c}{\underline{Structure-Recovery\&PSB}\\ShapeNet}\\
    \hline
    LeNet&0.27&0.35&0.21\\
    \hline
    L2-Net&0.23 & 0.40&0.26\\
    \hline
    HardNet&0.24 & 0.39&0.27\\
    \hline
    SOSNet& 0.24& 0.40&0.26\\
    \hline
    AlexNet-VP&0.26& 0.39&0.26\\
    \hline
    \sysName&\textbf{0.55}&\textbf{0.49}&\textbf{0.54}\\
    \hline
    \end{tabular}
       }
    \end{center}
   \caption{Cross-dataset (chair) performance of different methods in the sketch correspondence task. The training sets are labeled with underlines.
   }
   \label{table:tabcrossdatasetcross}
\end{table}
\begin{table}
   \begin{center}
       \resizebox{0.505\textwidth}{!}{
    \begin{tabular}{|l|c|c|c|}
    \hline
    Methods&\tabincell{c}{\underline{Structure-Recovery\&ShapeNet}\\PSB} & \tabincell{c}{\underline{PSB\&ShapeNet}\\Structure-Recovery} &\tabincell{c}{\underline{Structure-Recovery\&PSB}\\ShapeNet}\\
    \hline
    LeNet&0.07&0.01&0.04\\
    \hline
    L2-Net&0.16 & 0.35&0.13\\
    \hline
    HardNet&0.19 & 0.32&0.14\\
    \hline
    SOSNet& 0.17& 0.35&0.15\\
    \hline
    AlexNet-VP&0.54& 0.40&0.24\\
    \hline
    \sysName&\textbf{0.73}&\textbf{0.64}&\textbf{0.54}\\
    \hline
    \end{tabular}
       }
    \end{center}
   \caption{Cross-dataset (chair) performance of different methods in the pixel-wise retrieval task. The training sets are labeled with underlines.
   }
   \label{table:tabcrossdatasetretrival}
\end{table}
\subsubsection{View Disparity}
To further show the robustness of different learned descriptors against the degree of view disparity,
given the same input testing pixels in one sketch,
we visualize how the quality of correspondence inference changes with the increasing view disparity. As shown in Figure \ref{fig:Figure viewDisparity}, \sysName~shows a more stable performance of
correspondence inference than the competitors.
Please note that the ground-truth corresponding pixels might become invisible in certain views, and all the learned descriptors could not distinguish the visibility of corresponding pixels. Nevertheless, \sysName~still produces the most reasonable results.

\begin{figure*}[t]
    \begin{center}
    \includegraphics[width=0.9\linewidth]{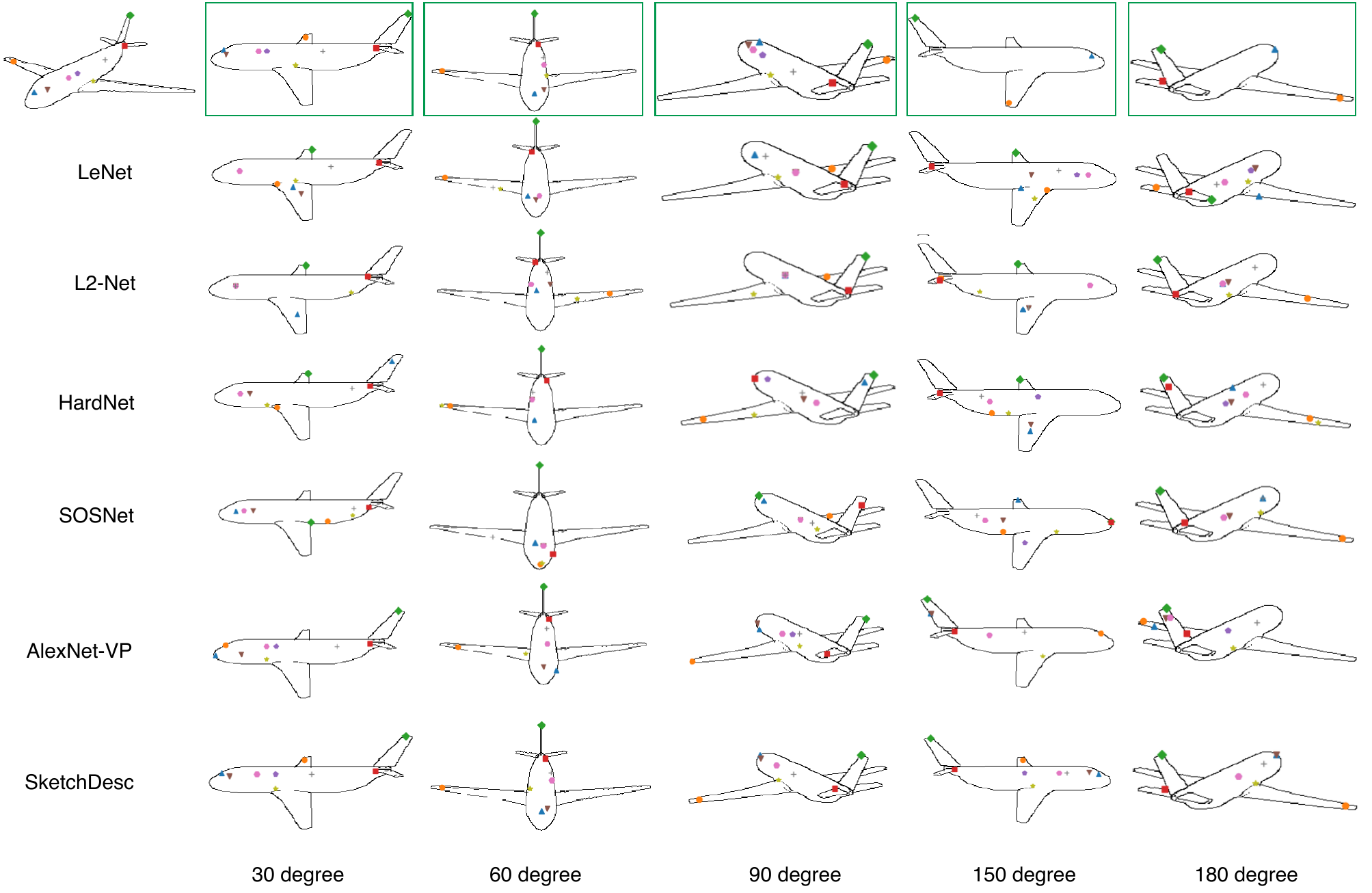}
    \end{center}
       \caption{Performance of different methods with increasing view disparity (30, 60, 90, 150, and 180 degrees). Given some anchor pixels on the sketched object at top-left, we show the corresponding pixels computed by the different methods. The ground-truth correspondences are labeled with green boxes.}
    \label{fig:Figure viewDisparity}
\end{figure*}

\subsection{Ablation Study}
\label{section: Ablation Study}
In this subsection, we validate the effectiveness of the key components of our method with ablation studies.

\textit{Multi-scale Strategy.}
The designed multi-scale patch-based representation  ($32 \times 32, 64 \times 64, 128 \times 128, 256 \times 256$) plays an essential role in our method. We first show how different scales can influence the performance of the learned descriptors. We test our network with increasingly more scales
and evaluate its performance on the pixel-wise retrieval task. We employ the average MAP (Mean Average Precision) metric over the whole dataset.
Quantitative results are shown in Table \ref{table:tabscales}.
We can see that the features from the larger scales are more discriminative than the features from smaller scales by ablating one of the multiple scales in Table \ref{table:tabscales} (rows 5 - 8).
As we remove the larger scale, the poorer performance of \sysName~is witnessed on all three datasets.
Multiple scales are better than a single scale.
A representative visual comparison is shown in Figure \ref{fig:Figure ablationStudy}.
It is found that as larger scales are involved, the ambiguous regions (bright yellow regions) on the feet, legs and backs of the camel are gradually rejected.
In other words, with the multi-scale patches as inputs, our network can enjoy not only a more precise local perception but also a global perspective.

\begin{figure}[t]
    \begin{center}
    \includegraphics[width=1\linewidth]{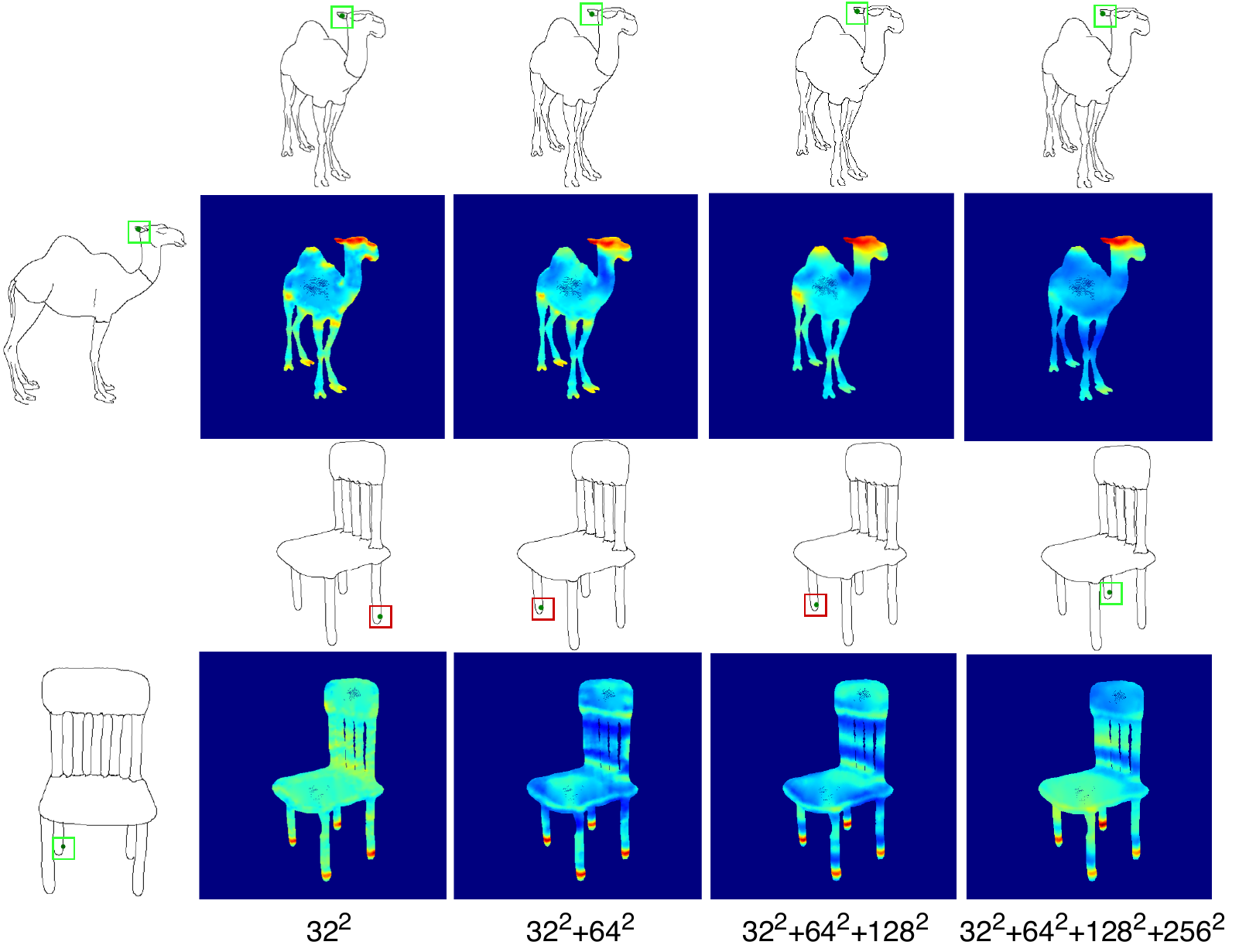}
    \end{center}
       \caption{Visualization of different multi-scale choices. The distance maps show the distances from the highlighted point in the left sketch to all the pixels in the other sketch.
       }
    \label{fig:Figure ablationStudy}
\end{figure}

\begin{table}
    \begin{center}
    \resizebox{0.89\columnwidth}{!}{
    \begin{tabular}{|l|c|c|c|}
    \hline
    Different scales &Structure-Recovery&PSB&ShapeNet\\
    \hline
    $32^2$&0.47&0.31&0.20\\
    \hline
    $64^2$&0.64&0.44&0.45\\
    \hline
    $128^2$&0.75&0.65&0.61\\
    \hline
    $256^2$&0.76 & 0.66 & 0.62 \\
    \hline
  $64^2+128^2+256^2$&0.81&0.69&0.66\\
    \hline
  $32^2+128^2+256^2$&0.81&0.67&0.65\\
    \hline
    $32^2+64^2+256^2$&0.79&0.68&0.64\\
    \hline
   $32^2+64^2+128^2$ &0.77&0.67&0.61\\
    \hline
      $32^2+64^2$&0.70&0.57&0.48\\
    \hline
   $32^2+64^2+128^2$&0.77&0.67&0.61\\
    \hline 
   $32^2+64^2+128^2+256^2$&\textbf{0.82}&\textbf{0.73}&\textbf{0.66}\\
    \hline
    \end{tabular}
       }
    \end{center}
    \caption{The performance of using different scale combinations as inputs to \sysName-Net in the pixel-wise retrieval task.
    }
    \label{table:tabscales}
\end{table}

\textit{Shared Weights.}
In our method, the multi-scale patches are processed by a shared-weight scheme. To verify its effectiveness, 
we perform a comparison on the pixel-wise retrieval task with an unshared-weight network structure.
The comparison results are reported in Table \ref{table:tabloss}. It is found that the shared-weight structure in our network achieves higher accuracy. The improvement is even more significant on PSB. The results confirm a similar design choice of shared-weight structure used in existing studies like  \cite{huang2018learning} and \cite{navarro2019sketchzooms}.

\begin{table}
    \begin{center}
        \resizebox{0.408\textwidth}{!}{
     \begin{tabular}{|l|c|c|c|}
     \hline
     &Structure-Recovery&PSB&ShapeNet\\
     \hline
     W/o  shared weights&0.80&0.65&0.63\\
     \hline
     W/ shared weights&\textbf{0.82}&\textbf{0.73}&\textbf{0.66}\\
     \hline
     \end{tabular}
    }
     \end{center}
    \caption{The performance of our method with or without shared-weights.
    }
    \label{table:tabloss}
\end{table}

\textit{Training Data Generation.}
We evaluate the performance of two patch sampling mechanisms on the task of pixel-wise retrieval: OR-sampling and AND-sampling (Section \ref{subsection:Data Preparation}). It can be found from Table \ref{table:tabdatapre} that OR-sampling achieves a significantly better performance. This is because the OR-sampling leads to a significantly larger dataset for training our network.

\begin{table}
    \begin{center}
        \resizebox{0.42\textwidth}{!}{
     \begin{tabular}{|l|c|c|c|}
     \hline
    Sampling Mechanism&Structure-Recovery&PSB&ShapeNet\\
     \hline
     {AND-sampling}
     &0.79&0.68&0.59\\
     \hline
     {OR-sampling}
     &\textbf{0.82}&\textbf{0.73}&\textbf{0.66}\\
     \hline
     \end{tabular}
        }
     \end{center}
    \caption{The performance of two sampling mechanisms for data preparation on the task of multi-view pixel-wise retrieval. Note that different sampling mechanisms are only used to generate the training data.
    }
    \label{table:tabdatapre}
\end{table}

\subsection{Limitations and Discussions}
\label{subsection: Limitation}
Our method has the following limitations. First, although our method could generalize well to unseen sketches or even hand-drawn sketches of the same objects, when the viewpoint differs from the examples in the training set drastically, our method could fail. This is a common generalization problem for any learning-based methods. Increasing the training data could help yet in the cost of additional training burden. 
We use a rather simple method to sample viewpoints for preparing the training data. A more careful view selection might be made by adopting best-view selection methods \cite{lee2005mesh}.
Additionally, our method is currently designed for multi-view correspondences of rigid objects. If the object undergoes articulation or non-rigid deformations (e.g. people dancing), our method may not perform well. We consider this as an intriguing future work to explore.

\section{Applications}

\subsection{Sketch Segmentation Transfer}

Sketch segmentation is a challenging task that demands plenty of human-labeled training data~\cite{Schneider:2016:CRF,li2018fast}. 
Noris et al.~\cite{noris2012smart} proposed a scribble-based UI for user-guided segmentation of sketchy drawings, which still requires substantial human efforts.
With \sysName, we show that segmentation labels can be easily transferred across multi-view sketches (Figure \ref{fig:Figure ApplicationSegTransfer}).
Specifically, we use \sysName~to produce 128-d descriptors for every point on the sketches.
With the correspondences established by the descriptors, we transfer the segmentation from one labeled sketch to other views.
As shown in Figure \ref{fig:Figure ApplicationSegTransfer},
although there are some distortions in the hand-drawn sketches, we can still obtain reasonable segmentation results through transfer.

\begin{figure}
	\begin{center}
		\includegraphics[width=0.8\linewidth]{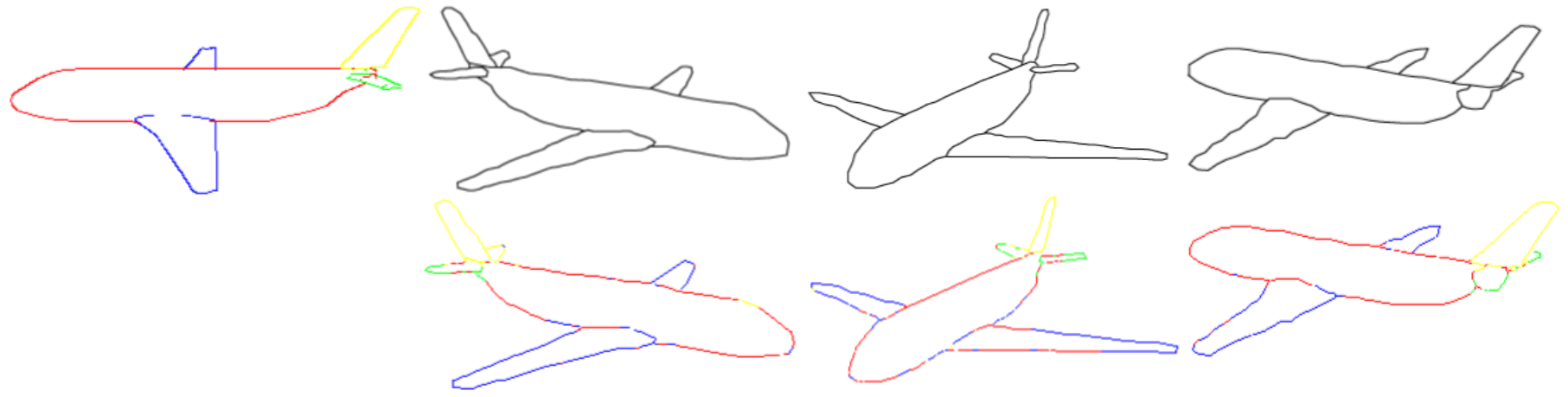}
	\end{center}
	\caption{Sketch segmentation transfer. The top row are the inputs: one segmented sketch and several unlabeled sketches. The bottom row are the outputs: sketches with point-wise labels after graph-cut postprocessing. With \sysName, we can transfer the labels among multi-view sketches with the computed correspondence.}
	\label{fig:Figure ApplicationSegTransfer}
\end{figure}

\begin{figure}
	\begin{center}
		\includegraphics[width=0.94\linewidth]{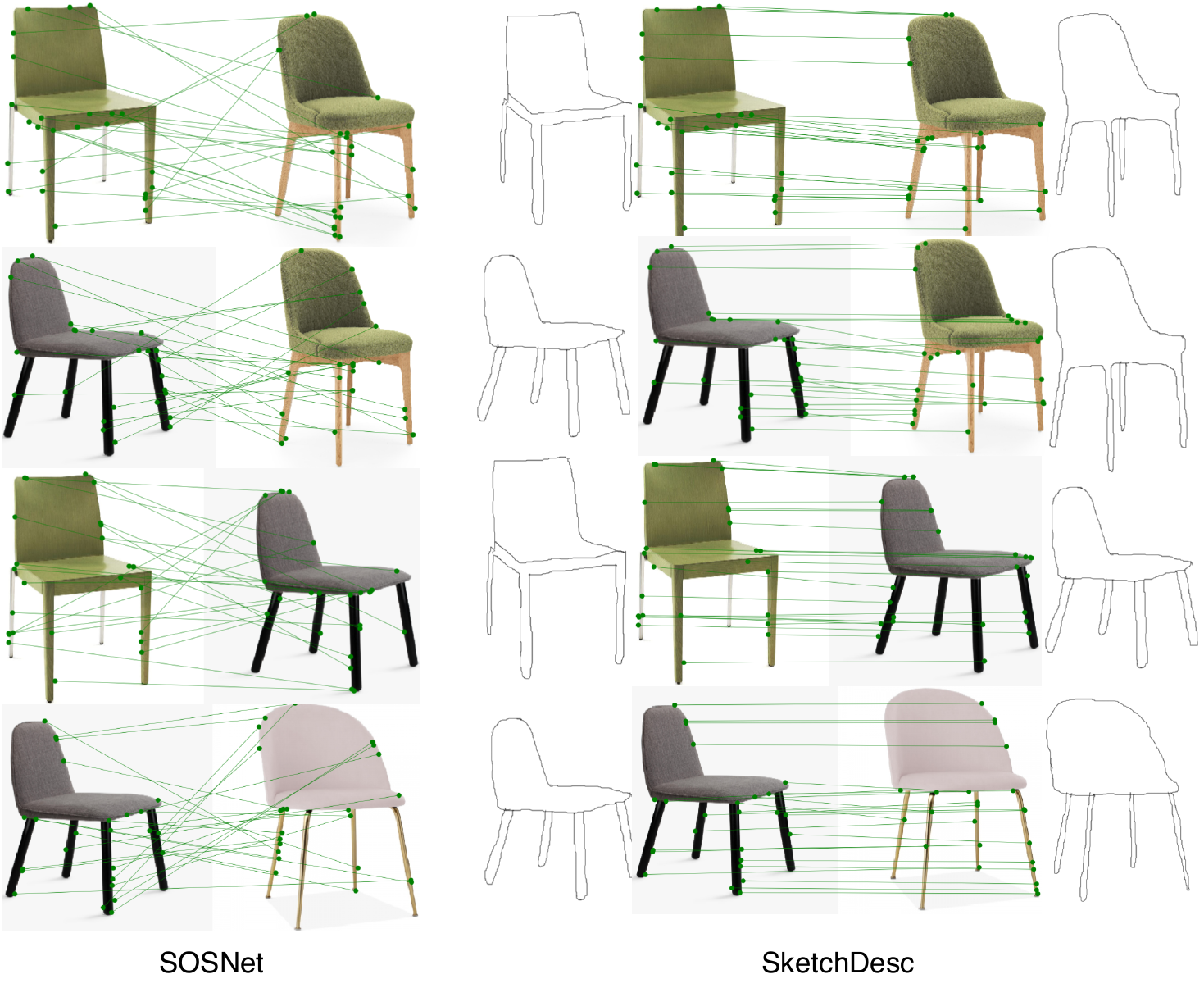}
	\end{center}
	\caption{Correspondence matching among multi-view images of different objects.}
	\label{fig:Figure ApplicationInterCorres}
\end{figure}

\subsection{Multi-view Image-based Correspondence}
Image-based descriptors \cite{tian2017l2,mishchuk2017working,tian2019sosnet} mostly rely on the information of textures in patches to build the correspondence among multi-view images. If the input images lack discriminative texture details (Figure \ref{fig:Figure ApplicationInterCorres}), the image-based methods (e.g. the state-of-art SOSNet \cite{tian2019sosnet}) may fail to extract robust descriptors with their single-scale input. Here we show that our sketch-based descriptor \sysName~can also be extended to this correspondence establishment among multi-view photos (Figure \ref{fig:Figure ApplicationInterCorres}) under such challenging situations even with no textures.
Here we use edge maps as a proxy, that is, we first convert a photo to its edge map format and then apply \sysName~to obtain local descriptors for matching corresponding points in different views. 
Figure \ref{fig:Figure ApplicationInterCorres} shows that \sysName~can infer a reasonable correspondence among multi-view images of different objects 
on mere edge maps, indicating the potential of our proposed \sysName~in the image domain. 
The reason why SketchDesc outperforms SOSNet
is because the multi-scale strategy in our method gives more confidence with the global and local perspectives.
This
further emphasizes the importance of th multi-scale idea in both the image and sketch domains.
On the other hand, while we believe \sysName~can assist existing photo correspondence techniques especially when images do not have rich textures, we do not claim that \sysName~is a general solution for photo correspondence.

\section{Conclusions}
\label{section: conclusions}
In this paper, we have introduced a deep learning based method for correspondence learning among multiple sketches of an object in different views. We have proposed a multi-branch network that encodes contexts from multi-scale patches with global and local perspectives to produce a novel descriptor for semantically measuring the distance of pixels in multi-view sketch images. The multi-branch and shared-weights designs help the network capture more feature information from all scales of sketch patches. 
Our data preparation method provides the ground truth effectively for training our multi-branch network. We believe the generated data can benefit other applications.
Both qualitative and quantitative experiments show that our learned descriptor is more effective than the existing learning-based descriptors.
In the future, it would be interesting to exploit more neighboring information and learn the per-point features in a joint manner.

\ifCLASSOPTIONcaptionsoff
  \newpage
\fi
\bibliographystyle{IEEEtran}
\bibliography{egbib}

\begin{IEEEbiography}[{\includegraphics[width=1in,height=1.25in,clip,keepaspectratio]{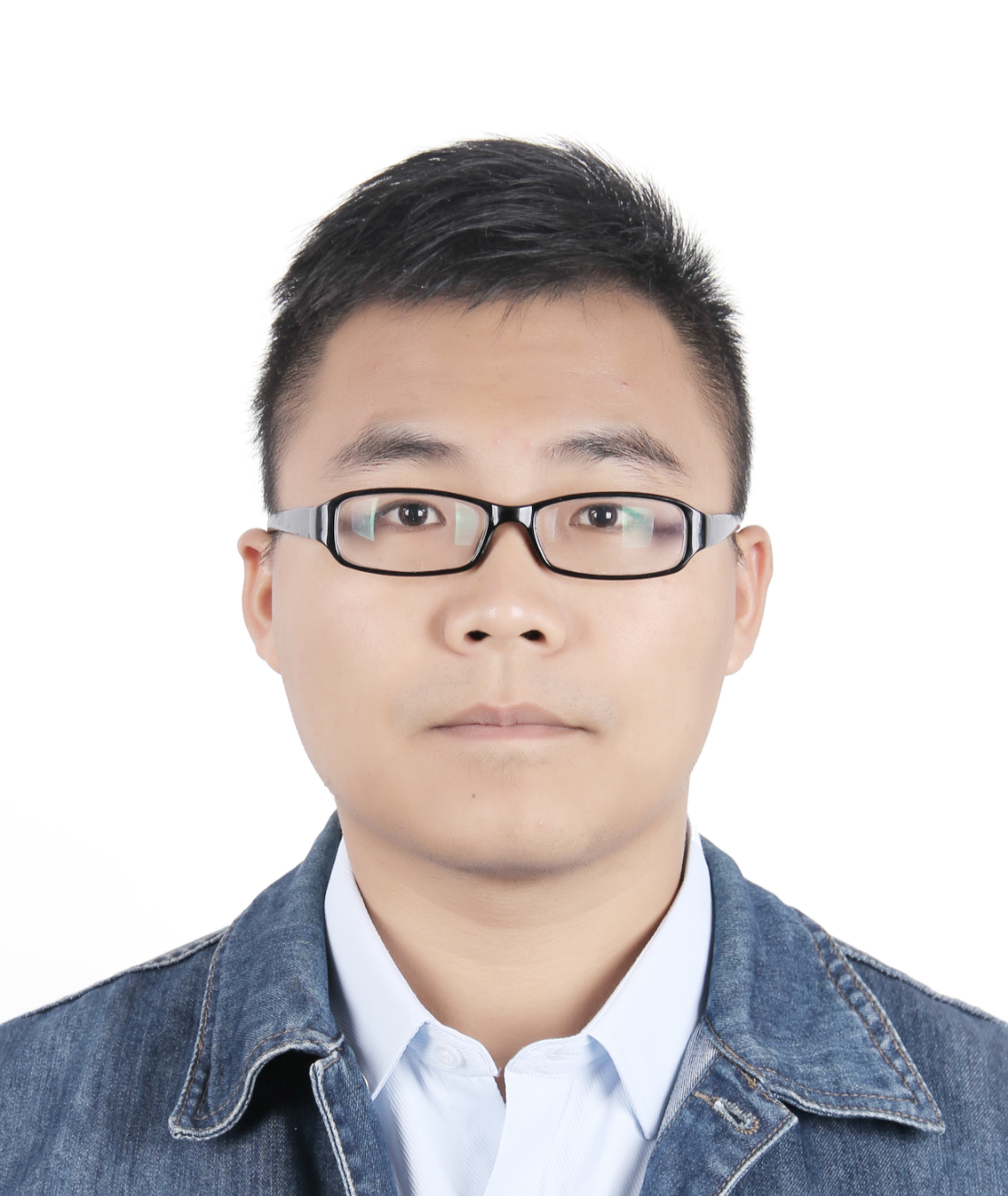}}]{Deng Yu}
is pursuing the Ph.D. degree at
the School of Creative Media, City University of Hong Kong. He received the B.Eng. degree and the Master degree in computer science and technology from China University of Petroleum (East China). His research interests include computer graphics and data-driven techniques.
\end{IEEEbiography}
\vfill
\begin{IEEEbiography}[{\includegraphics[width=1in,height=1.25in,clip,keepaspectratio]{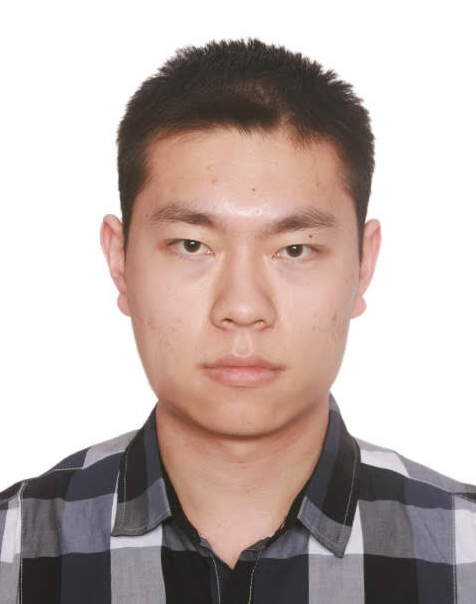}}]{Lei Li}
  is working toward the Ph.D. degree at
  the Department of Computer Science and Engineering, Hong Kong University of Science and Technology. He received the B.Eng. degree in software engineering from Shandong University. His research interests include computer graphics and data-driven techniques.
\end{IEEEbiography}
\vfill
\begin{IEEEbiography}[{\includegraphics[width=1in,height=1.25in,clip,keepaspectratio]{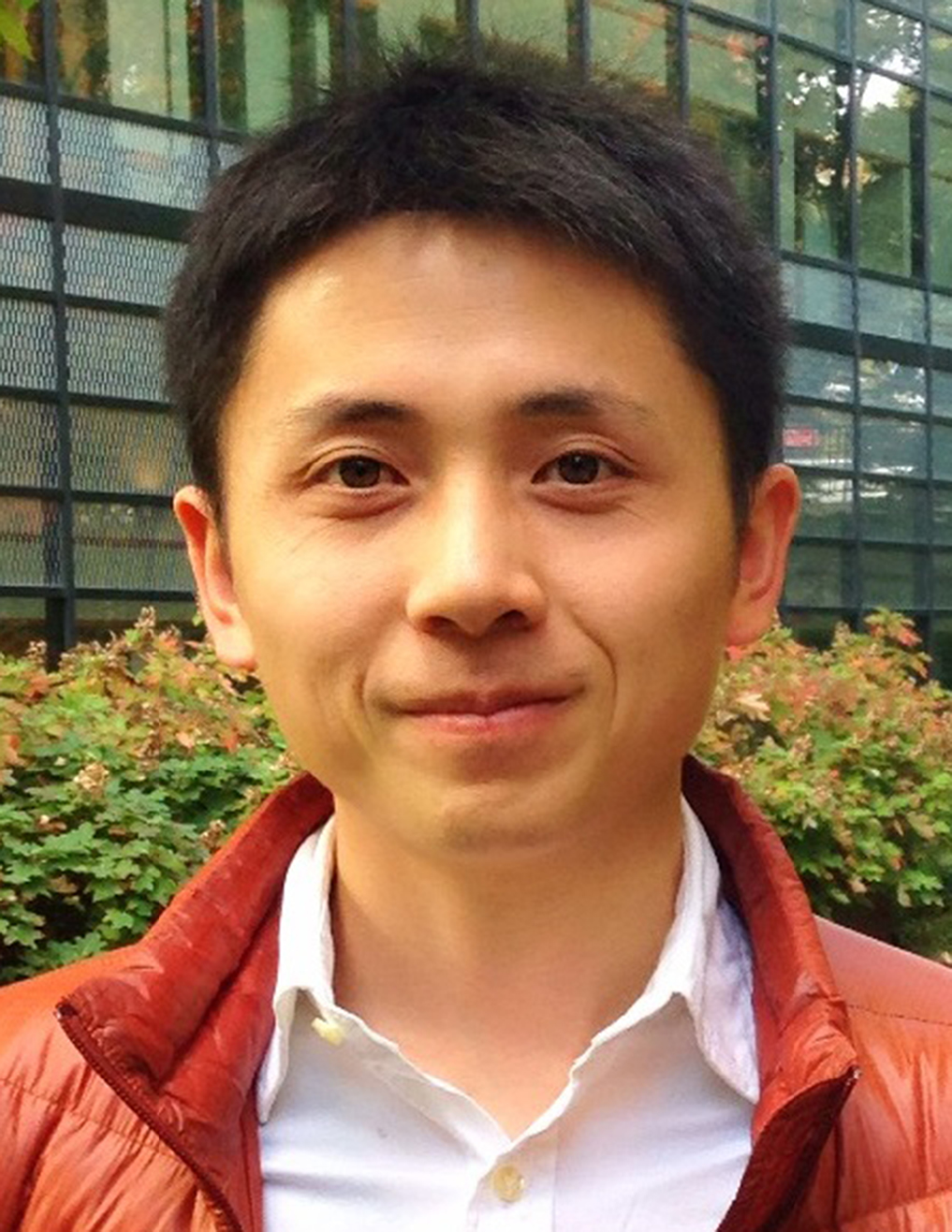}}]{Youyi Zheng}
is a Researcher at the State
Key Lab of CAD\&CG, College of Computer Science, Zhejiang University. He obtained his Ph.D. from the Department of Computer Science and Engineering at Hong Kong University of Science and Technology, and his M.Sc. and B.Sc. degrees in Mathematics, both from Zhejiang University. His research interests include geometric modeling, imaging, and human-computer interaction.
\end{IEEEbiography}
\vfill
\begin{IEEEbiography}[{\includegraphics[width=1in,height=1.25in,clip,keepaspectratio]{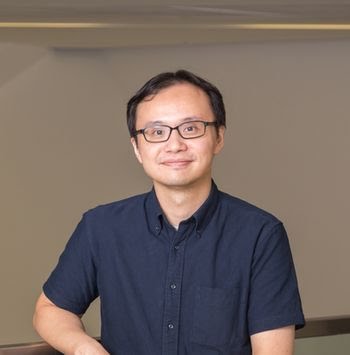}}]{Manfred Lau}
is an Assistant Professor in the School of Creative Media at the City University of Hong Kong. His research interests are in computer graphics, human-computer interaction, and digital fabrication. His recent research in the perception of 3D shapes uses crowdsourcing and learning methods for studying human perceptual notions of 3D shapes. He was previously an assistant professor in the School of Computing and Communications at Lancaster University in the UK, and a post-doc researcher in Tokyo at the Japan Science and Technology Agency - Igarashi Design Interface Project. He received his Ph.D. degree in Computer Science from Carnegie Mellon University, and his B.Sc. degree in Computer Science from Yale University. He has served in the program committees of the major graphics conferences including Siggraph Asia.
\end{IEEEbiography}
\vfill
\begin{IEEEbiography}[{\includegraphics[width=1in,height=1.25in,clip,keepaspectratio]{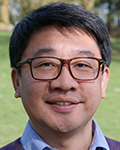}}]{Yi-Zhe Song} is a Reader of Computer Vision and Machine Learning at the
Centre for Vision Speech and Signal Processing (CVSSP), where he
directs the SketchX lab. Previously, he was a Senior Lecturer at the Queen Mary
University of London, and a Research and Teaching Fellow at the University of Bath. He obtained his PhD in 2008 on Computer Vision and Machine Learning from the University of Bath, and received a Best Dissertation Award from his MSc degree at the University of Cambridge in 2004, after getting a First Class Honours degree from the University of Bath in 2003. He is a Senior Member of IEEE, and a Fellow of the Higher Education Academy. He is a full member of the review college of the Engineering and Physical Sciences Research Council (EPSRC), the UK's main agency for funding research in engineering and the physical sciences, and serves as an expert reviewer for the Czech National Science Foundation.
\end{IEEEbiography}
\vfill
\begin{IEEEbiography}[{\includegraphics[width=1in,height=1.25in,clip,keepaspectratio]{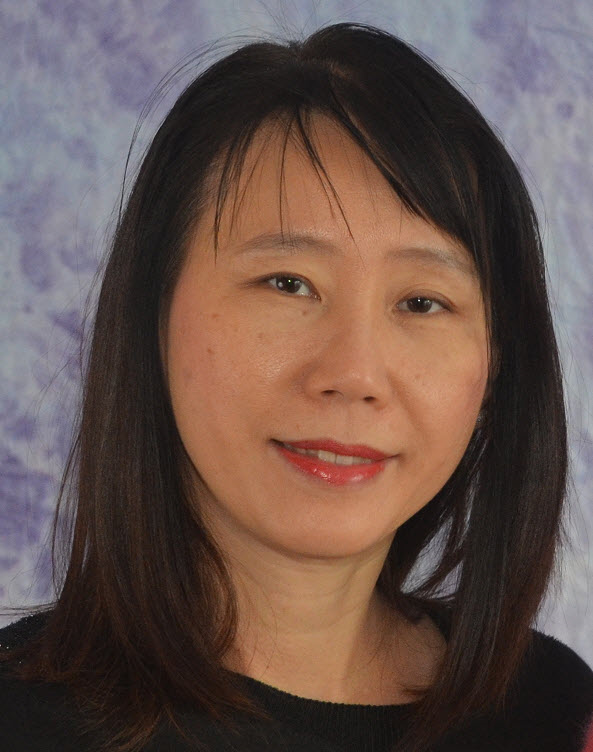}}]{Chiew-Lan Tai}
is a Professor at the Department
of Computer Science and Engineering, Hong
Kong University of Science and Technology. She
received the B.Sc. degree in mathematics from
University of Malaya, the M.Sc. degree in computer and information sciences from National University of Singapore, and the D.Sc. degree in information science from the University of Tokyo. Her research interests include geometry processing, computer graphics, and interaction techniques.
\end{IEEEbiography}
\vfill
\begin{IEEEbiography}[{\includegraphics[width=1in,height=1.25in,clip,keepaspectratio]{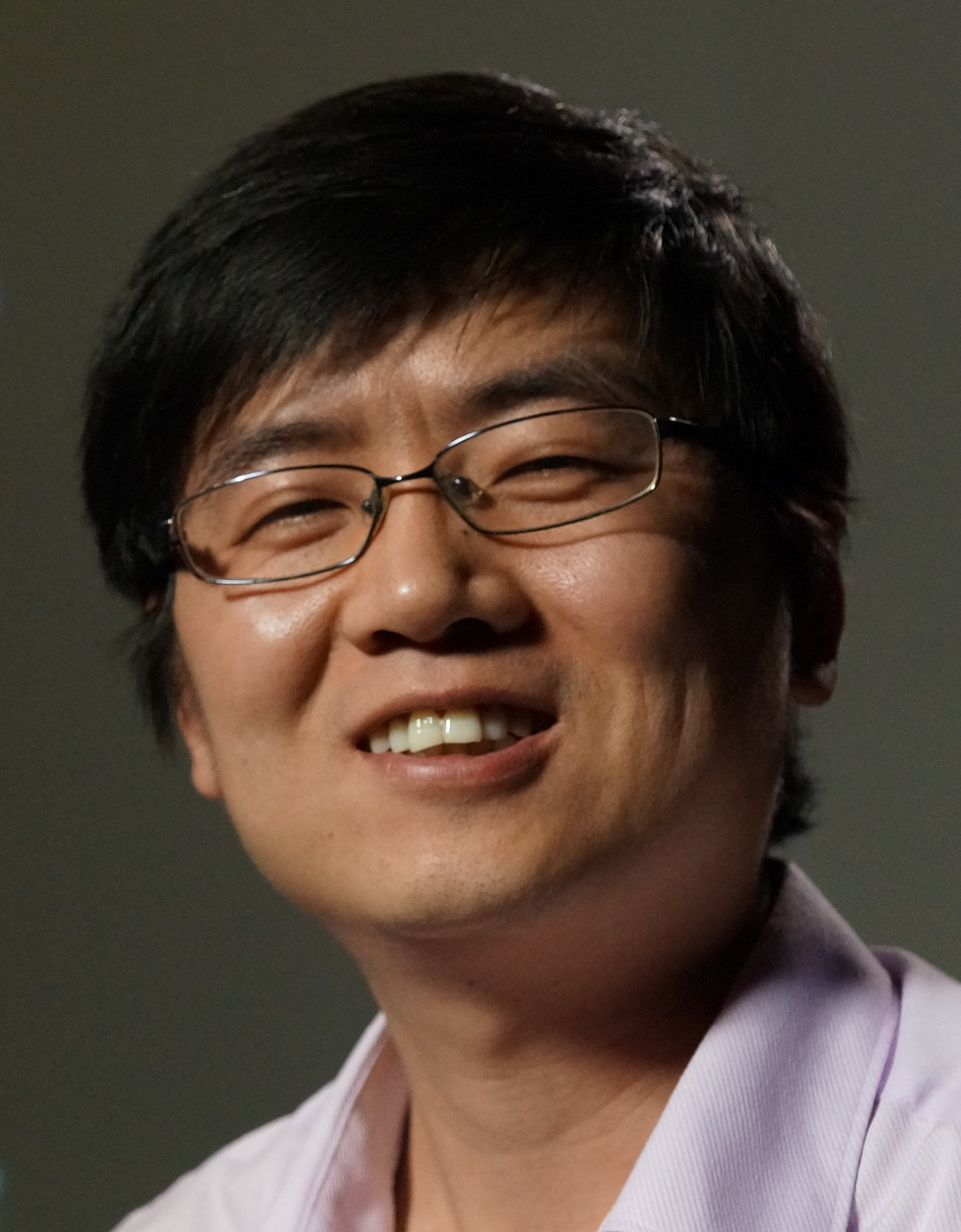}}]{Hongbo Fu}
is a Professor with the School of
Creative Media, City University of Hong Kong.
He received the B.S. degree in information sciences from Peking University, and the Ph.D.
degree in computer science from Hong Kong
University of Science and Technology. He has
served as an Associate Editor of The Visual
Computer, Computers \& Graphics, and Computer
Graphics Forum. His primary research interests
include computer graphics and human
computer interaction.
\end{IEEEbiography}
\vfill

\end{document}


 \title{Supplemental Materials\\SketchDesc: Learning Local Sketch Descriptors for Multi-view Correspondence}
\maketitle

\begin{figure*}[hb]
   \centering
   \includegraphics[width=0.9\linewidth]{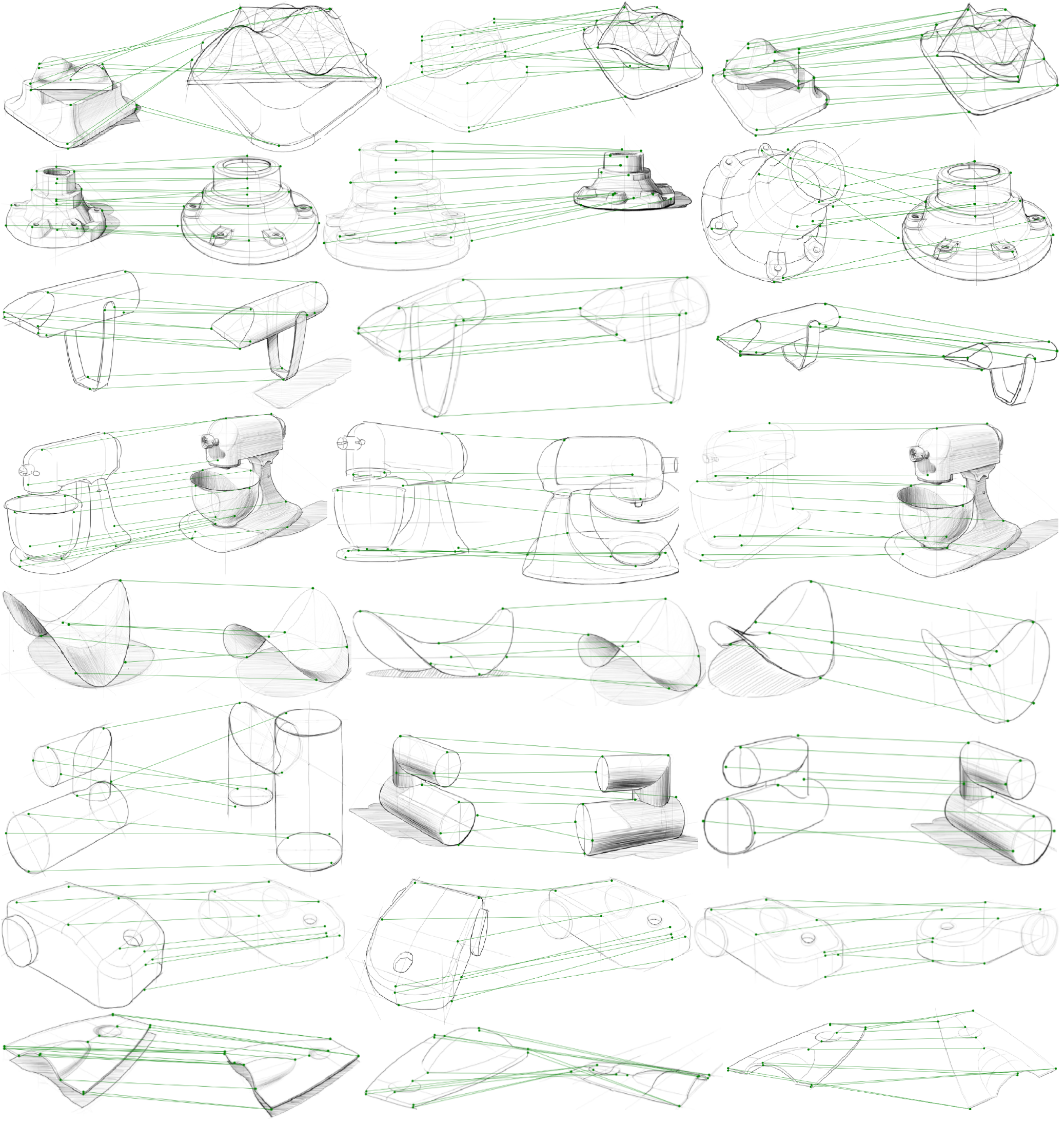}
   \caption{Correspondence of multi-view sketches built by \sysName~descriptors. All of the multi-view sketches are from the OpenSketch dataset.}
   \label{fig:appendixOpenSketch}
\end{figure*}

\begin{figure*}
   \centering
   \includegraphics[width=0.94\linewidth]{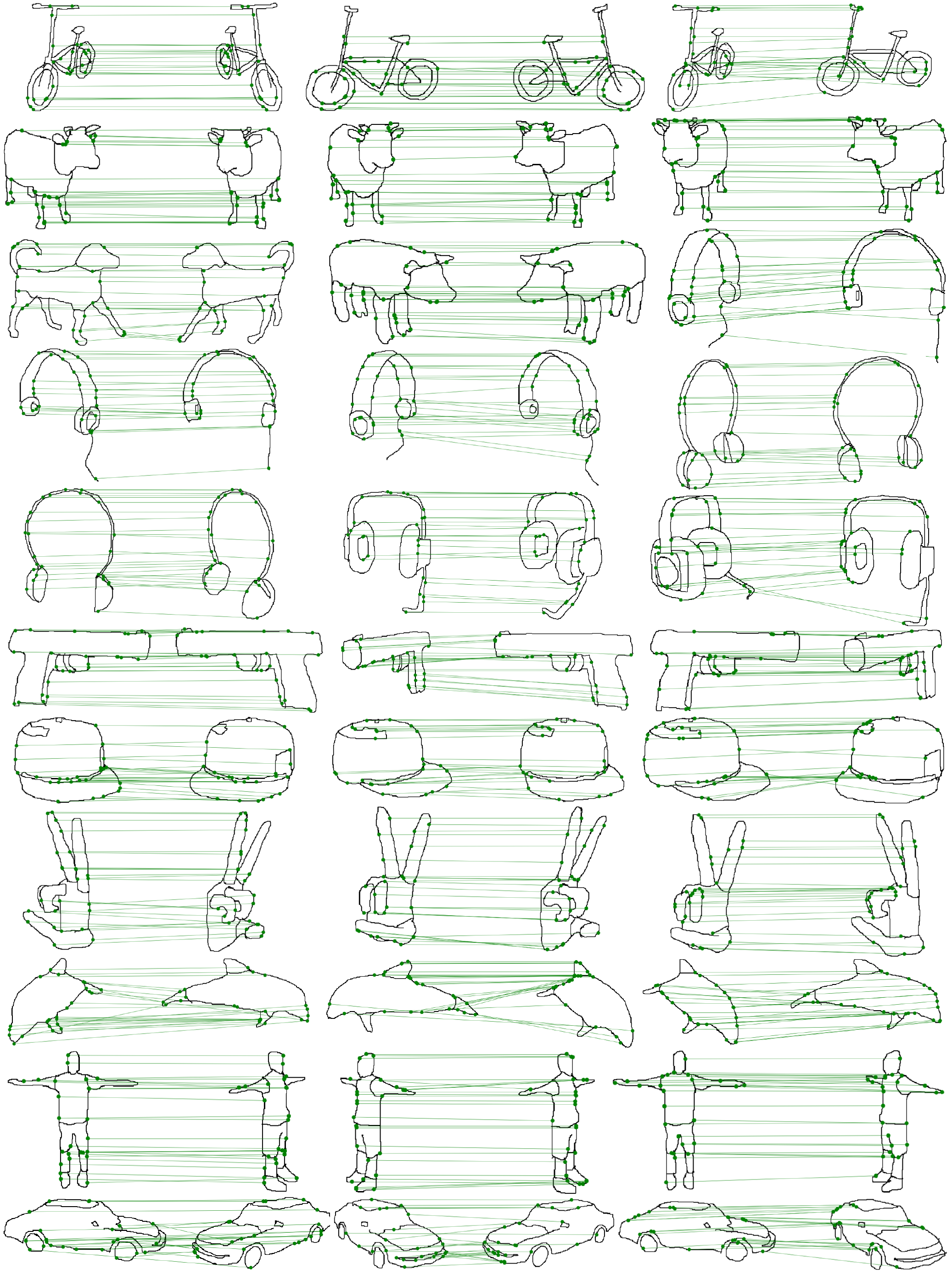}
   \caption{Correspondence of multi-view sketches built by \sysName~descriptors. All of these multi-view sketches are created by volunteers.}
   \label{fig:appendixHumanSketch}
\end{figure*}

\label{appendix B}
\begin{figure*}
   \centering
   \includegraphics[width=1\linewidth]{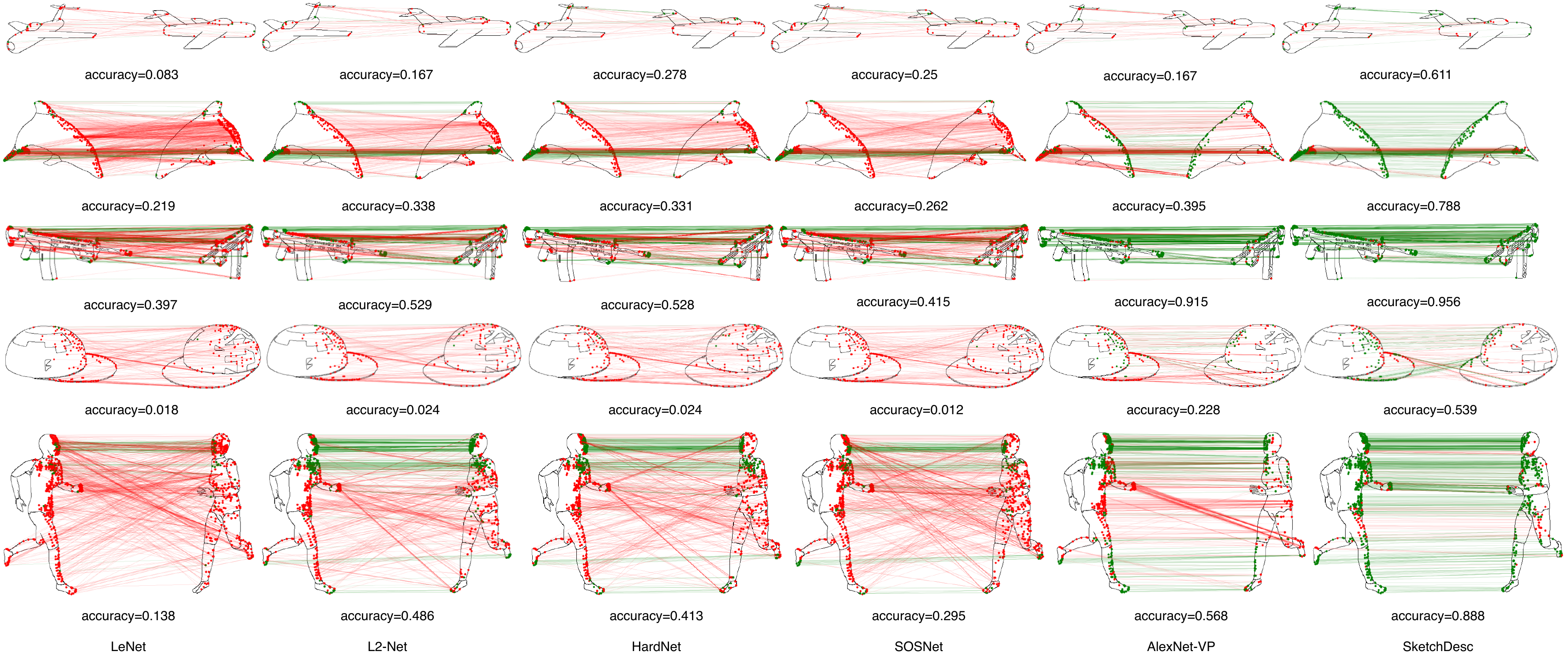}
   \caption{Sketch correspondence for multi-view sketches {(synthesized)}. The red lines indicate the failed matching and the green lines show the matching correspondences among multi-view sketches.}
   \label{fig:appendixFig1}
\end{figure*}
